\documentclass[10pt,twocolumn,letterpaper]{article}

\usepackage[pagenumbers]{cvpr} 

\definecolor{cvprblue}{rgb}{0.21,0.49,0.74}
\usepackage[pagebackref,breaklinks,colorlinks,allcolors=cvprblue]{hyperref}

\def\paperID{30486} 
\def\confName{CVPR}
\def\confYear{2026}

\title{MU-GeNeRF: Multi-view Uncertainty-guided Generalizable Neural Radiance Fields for Distractor-aware Scene}

\author{
    Wenjie Mu$^{1,}$\thanks{Equal contribution} \quad 
    Zhan Li$^{1,*}$ \quad 
    Chuanzhou Su$^{1}$ \quad 
    Xuanyi Shen$^{1}$ \\
    Ziniu Liu$^{1}$ \quad 
    Fan Lu$^{1}$ \quad 
    Yujian Mo$^{1}$ \quad 
    Junqiao Zhao$^{1,2,}$\thanks{Corresponding author} \\ 
    Tiantian Feng$^{3}$ \quad 
    Chen Ye$^{1,2}$ \quad 
    Guang Chen$^{1,2,4}$ \\[2mm]
    $^{1}$School of Computer Science and Technology, Tongji University, Shanghai, China \\
    $^{2}$MOE Key Lab of Embedded System and Service Computing, Tongji University, Shanghai, China \\
    $^{3}$College of Surveying and Geographic Informatics, Tongji University, Shanghai, China \\
    $^{4}$Shanghai Innovation Institute
}

\begin{document}
\maketitle

\begin{abstract}
Generalizable Neural Radiance Fields (GeNeRFs) enable high-quality scene reconstruction from sparse views and can generalize to unseen scenes. However, in real-world settings, transient distractors break cross-view structural consistency, corrupting supervision and degrading reconstruction quality. Existing distractor-free NeRF methods rely on per-scene optimization and estimate uncertainty from per-view reconstruction errors, which are not reliable for GeNeRFs and often misjudge inconsistent static structures as distractors. 
To this end, we propose MU-GeNeRF, a Multi-view Uncertainty-guided distractor-aware GeNeRF framework designed to alleviate GeNeRF's robust modeling challenges in the presence of transient distractions. 
We decompose distractor awareness into two complementary uncertainty components: Source-view Uncertainty, which captures structural discrepancies across source views caused by viewpoint changes or dynamic factors; 
and Target-view Uncertainty, which detects observation anomalies in the target image induced by transient distractors.
These two uncertainties address distinct error sources and are combined through a heteroscedastic reconstruction loss, which guides the model to adaptively modulate supervision, enabling more robust distractor suppression and geometric modeling.
Extensive experiments show that our method not only surpasses existing GeNeRFs but also achieves performance comparable to scene-specific distractor-free NeRFs.
Project page: \url{https://github.com/Yanyilucas/MU-GeNeRF}.

\end{abstract}

\vspace{-3pt}
\section{Introduction}
\label{sec:intro}
Neural Radiance Fields (NeRF) \cite{mildenhall2021nerf} is a novel view synthesis technique based on implicit scene representation, with broad applications in virtual reality \cite{deng2022fov,dai2025go,xu2023vr}, 3D reconstruction \cite{neupane2025high,wang2024mp,zhang2024supernerf} and autonomous driving \cite{cao2024lightning,hu2024pc,tonderski2024neurad}. To enhance its adaptability to open-world environments, recent studies have introduced Generalizable Neural Radiance Fields (GeNeRF) \cite{yu2021pixelnerf}, which learn a scene-agnostic multi-view aggregation prior to achieve cross-scene implicit 3D reconstruction. 
Both NeRF and GeNeRF follow the assumption that scene geometry and lighting remain static during data capture \cite{gao2022nerf,su2024boostmvsnerfs,chen2025explicit,lin2022efficient,li2023improved,yang2023contranerf,cong2023enhancing}. However, transient distractors such as shadow variations and dynamic objects are common in real-world environments, disrupting structural consistency across views and introducing erroneous supervision, ultimately leading to reconstruction distortions \cite{debbagh2023neural,yao2024neural}.

\begin{figure}[t!]
\centering
\includegraphics[width=0.90\linewidth]{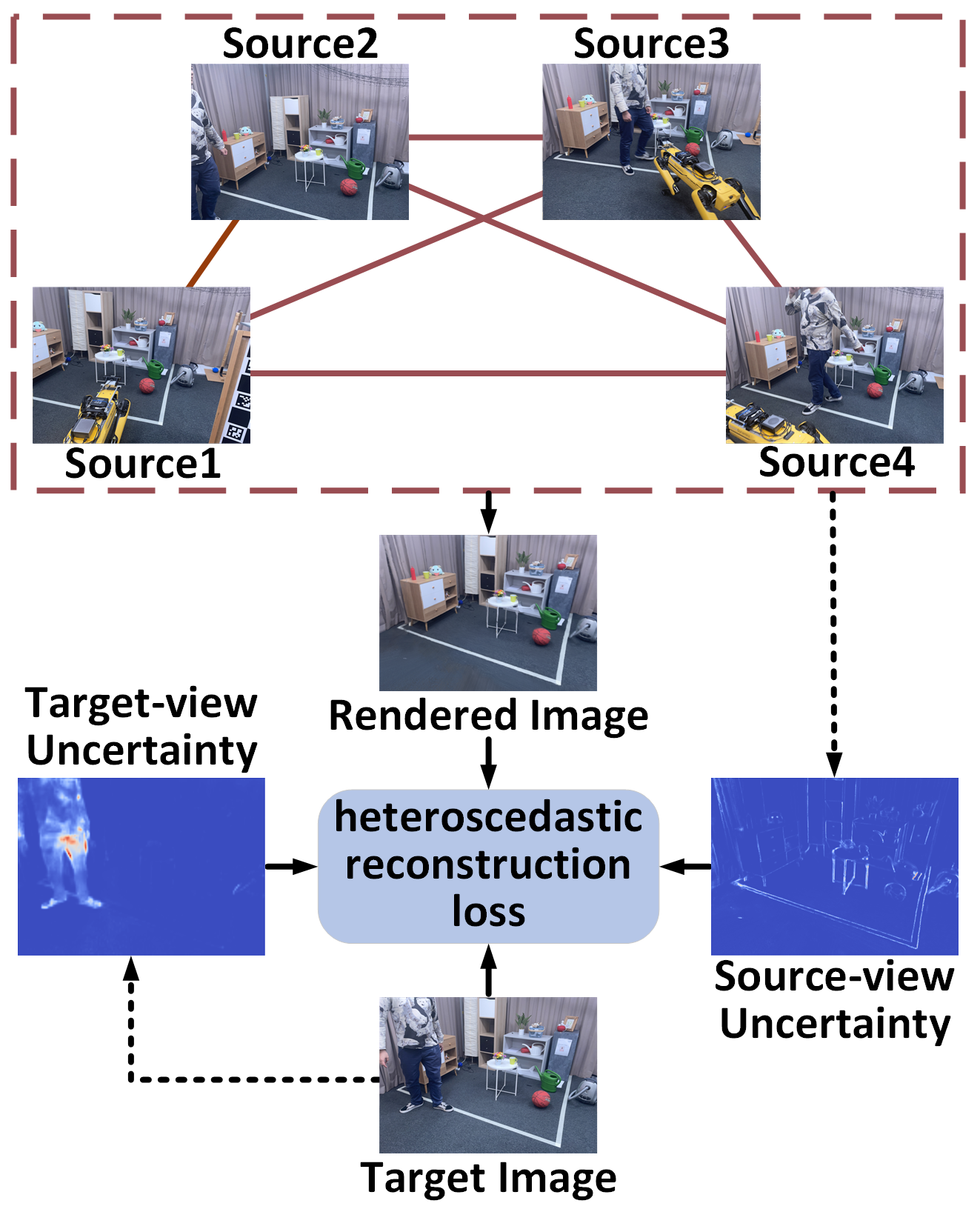}
\vspace{-0.5em}
\setlength{\belowcaptionskip}{-12.2pt} 
\caption {\textbf{Proposed Multi-view Uncertainty-guided distractor-aware methods.} Two complementary uncertainties are estimated to support stable training in dynamic scenes. Source-view Uncertainty captures structural inconsistencies across source views arising from occlusion or dynamic factors, while Target-view Uncertainty localizes the spatial distribution of distractors within the target view. Together, they synergistically form a heteroscedastic reconstruction loss that guides the model to adaptively modulate supervision and enhances the robustness of geometric modeling.}
\label{fig:Framework}
\end{figure}
To mitigate the impact of transient distractors on NeRF, some studies introduce uncertainty modeling to identify and remove them. For example, NeRF-W \cite{martin2021nerf} explicitly models distractors by introducing a transient embedding to estimate the uncertainty for each sampled point, separating them from the static scene. NeRF on-the-go \cite{ren2024nerf} treats transient distractors as high-uncertainty outliers during training and removes their influence by constructing a robust heteroscedastic loss. 
These methods are built upon NeRF’s per-scene optimization paradigm, which relies on sufficient training with multi-view inputs to establish consistency and estimate uncertainty from per-view reconstruction errors—essentially separating static structures from transient distractors by overfitting to a specific scene.

However, this paradigm is not transferable for generalization settings, as GeNeRF does not perform independent optimization for each scene, but instead learns a universal multi-view aggregation prior to synthesize the target view directly from sparse source views. As a result, reconstruction errors may stem from a variety of factors, arising both from transient distractors in the target view and structural inconsistencies across source views due to occlusion or dynamic factors. Indiscriminately utilizing all reconstruction errors to model distractors is highly prone to misjudging inconsistent static structures as distractors. This not only weakens uncertainty interpretability but also severely impairs geometric modeling accuracy.

To address this limitation, we propose a Multi-view Uncertainty-guided distractor-aware GeNeRF framework (MU-GeNeRF), which effectively suppresses cross-view structural inconsistencies and transient distractors during training, ultimately enhancing the robustness of geometric modeling.
As illustrated in Fig. \ref{fig:Framework}, we introduce two complementary forms of uncertainty: Source-view Uncertainty, modeled through a generalizable feed-forward process that captures structural discrepancies across source views caused by occlusions or dynamic factors; 
and Target-view Uncertainty, which identifies observation anomalies caused by transient distractors in the target image. 
Together, these uncertainties resolve the ambiguity inherent in attributing reconstruction errors in generalized settings.
Finally, we integrate these two uncertainties into a heteroscedastic reconstruction loss, guiding the model to adaptively modulate supervision to effectively suppress distractors, thereby providing a more stable training for geometric modeling. 
Extensive experiments on relevant datasets demonstrate the reliability and effectiveness of the proposed method, further extending the applicability of GeNeRF to dynamic real-world scenes.

\section{Related Work}
\vspace{-7pt}
\label{sec:Work}
\paragraph{Generalizable Neural Radiance Fields}
GeNeRF adopts a feed-forward design, directly modeling scene representations by aligning features with query coordinates, thus avoiding the cumbersome steps of single-scene optimization \cite{chibane2021stereo}. The success of such methods often relies on advanced network architectures for effectively aligning and aggregating multi-view information \cite{nguyen2023cascaded,du2023learning,ni2024colnerf,min2024entangled}. 
For instance, IBRNet \cite{wang2021ibrnet} uses a Transformer with cross-attention to aggregate multi-view features efficiently, boosting cross-view information interaction and representation; MVSNeRF \cite{chen2021mvsnerf} uses cost volume to enforce depth consistency across views, improving geometric accuracy; MuRF \cite{xu2024murf} employs 3D CNNs to process voxel data, capturing local spatial features via convolutions for fine-grained scene modeling.
However, these methods \cite{zhu2024caesarnerf,wang2022attention,wang2025golf,tanay2024global} generally rely on the assumption of static scenes and thus struggle to handle dynamic objects. Recent GeNeRF approaches address this limitation by incorporating temporal modeling \cite{liu2023robust,sun2024dyblurf,zhao2025dynamic} to represent object motion. For example, DynIBR \cite{li2023dynibar} aggregates features in ray space to account for motion and camera shifts; CTNeRF \cite{miao2024ctnerf} uses a temporal Transformer to handle changes in monocular videos. Unlike these approaches, which focus on modeling object motion, our method targets the detection and suppression of transient distractors present in the scene.

\begin{figure*}[t]
  \centering
  \includegraphics[width=1.0\textwidth]{sec/images/fig2.png}
  \vspace{-0.7em}
  \setlength{\belowcaptionskip}{-6pt} 
  \caption{\textbf{Overall architecture of our method.} Given multi-view inputs, the model projects sampled points from target-view rays onto source images $\{ I_n^s \}_{n=1}^N$ and feature maps $\{ F_n^s \}_{n=1}^N$, sampling color $\{ c_n^s \}_{n=1}^N$ and feature information $\{ f_n^s \}_{n=1}^N$. Combined with spatial information $x_k$ and $d_k$, these are fed into an feed-forward network $\mathcal{G}_\theta$ to predict each point’s color $c_k$ and source-view uncertainty $\beta_k^s$. Next, treating each prediction as a gaussian distribution, a Gaussian Mixture Model along the ray produces the rendered color $\hat{C}$ and source-view uncertainty $\beta^s$. Meanwhile, the target image $I^t$ is processed by a DINOv2 network to extract semantic features $F^t$, which are used by a decoder $\mathcal{F}_\theta$ to estimate a dense target-view uncertainty map $\beta^t$. Finally, the two uncertainties synergistically form a heteroscedastic reconstruction loss that adaptively modulates supervision to enhance the robustness of geometric modeling}
  \label{Fig:arch}
\end{figure*} 

\paragraph{Distractor-free NeRF}
Applying NeRF to real-world scenes inevitably faces transient distractors that disrupt cross-view consistency, causing blurred or distorted reconstructions. Existing solutions fall into two categories: one uses pre-trained semantic segmentation models to detect transient distractors and mask unreliable regions during training \cite{sun2022neural,schischka2024dynamon,chen2024nerf,turki2022mega,tancik2022block,rematas2022urban}, but these methods rely on prior knowledge and cannot identify unseen object classes. The other category identifies and removes transient distractions by introducing uncertainty estimation or explicit mask modeling \cite{chen2022hallucinated,li2023nerf,kassab2023refinedfields,yang2023cross,lee2023semantic,wang2025ie}.
For example, UP-NeRF \cite{kim2023up} estimates uncertainty for each sampled point and explicitly models transient distractors, thereby separating them from the static scene; NeRF on-the-go \cite{ren2024nerf} estimates uncertainty at the ray level and implicitly removes distractors by formulating a robust heteroscedastic loss.
However, both methods rely on NeRF’s per-view modeling, estimating uncertainty from single-view reconstruction errors and requiring dense views and scene-specific training. 
In contrast, we model uncertainty separately for target and source views to distinguish transient distractors from structural inconsistencies caused by viewpoint differences, enabling more accurate distraction suppression in GeNeRF.

\section{Preliminaries}	
\label{sec:Preliminaries}
\vspace{-3pt}
\subsection{Generalizable NeRF}
In a typical GeNeRF view synthesis pipeline, given $N$ source images $\{ I_n^s \}_{n=1}^N$ with known camera parameters, a shared image encoder extracts corresponding feature representations $\{ F_n^s \}_{n=1}^N$. The model then uniformly samples points along the ray $\{ \mathbf{r}(t) = \mathbf{o} + t\mathbf{d},\ t \geq 0 \}$ for each pixel in the target images $I^t$. For each sample point $t_k$, the projection function $\pi_n(\cdot)$ projects it onto the source image feature $F_n^s$ to sample the projected feature $f_n^s$:
\begin{equation}\label{eq:eq1}
f_n^{s}(x_k) = \chi\left(F_n^{s}, \pi_n(x_k)\right)
\end{equation}
where $\chi(\cdot)$ denote the bilinear interpolation function. Next, we feed the projected features $f_n^s$ along with spatial information $x_k$ and $d_k$ into the feed-forward network $\mathcal{G}_\theta$ to predict the color $c_k$ and $\rho_k$ density at each sampled point:
\begin{equation}\label{eq:eq2}
c_k, \rho_k = \mathcal{G}_\theta\left(f_n^{s}, x_k, d_k\right)
\end{equation}

Finally, the discrete colors and densities sampled along the ray are integrated via volume rendering to produce the predicted pixel color $\hat{C}(r)$.

\subsection{Uncertainty-Guided Distractor Suppression}
Based on the assumption that transient distractor regions are usually accompanied by higher uncertainty, pixel-level uncertainty is used to modulate the regional supervision in GeNeRF’s reconstruction loss, enabling adaptive suppression of distractors. Specifically, given the ground-truth color $C(r)$ and the predicted color $\hat{C}(r)$, the heteroscedastic reconstruction loss can be formulated as follows:
\begin{equation}\label{eq:eq3}
\mathcal{L}_{\text{uncer}} = \frac{\mathcal{L}_{\text{MSE}}(C(r), \hat{C}(r))}{2\beta^2(r)} + \lambda \log \beta(r)
\end{equation}
where $\lambda$ is a regularization coefficient. The first term adaptively adjusts the reconstruction error weights across pixels based on the predicted uncertainty $\beta(r)$ applying weaker supervision to high-uncertainty regions. The second term is a logarithmic regularization that constrains the range of $\beta(r)$, preventing it from diverging toward infinity. 

\vspace{-3pt}
\section{Method}	
Fig. \ref{Fig:arch} illustrates the overall framework of MU-GeNeRF. 
We introduce Source-view Uncertainty, modeled through a generalizable feed-forward process that captures geometric and appearance discrepancies across source views.
Meanwhile, target-view Uncertainty is estimated from semantic features to localize the spatial distribution of distractors in the target view.
Finally, both uncertainties are integrated into a heteroscedastic reconstruction loss, enabling precise perception and effective suppression of distractor regions, thus ensuring stable and consistent reconstruction.

\begin{figure}[t]
\centering
\setlength{\tabcolsep}{1.5pt}
\begin{tabular}{cccc}
    \begin{subfigure}{0.295\columnwidth}
    \includegraphics[width=\linewidth,height=0.6667\linewidth]{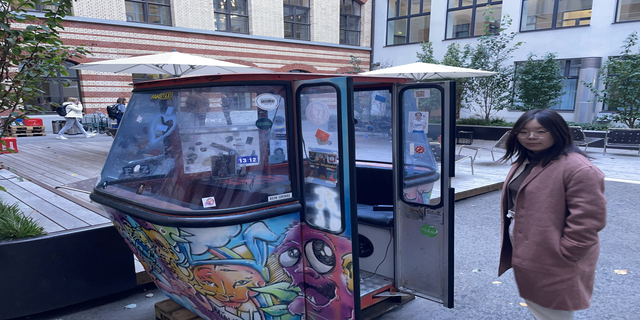}
    \caption{GT Image}
    \label{fig:3.1}
    \end{subfigure} &
    \begin{subfigure}{0.295\columnwidth}
    \includegraphics[width=\linewidth,height=0.6667\linewidth]{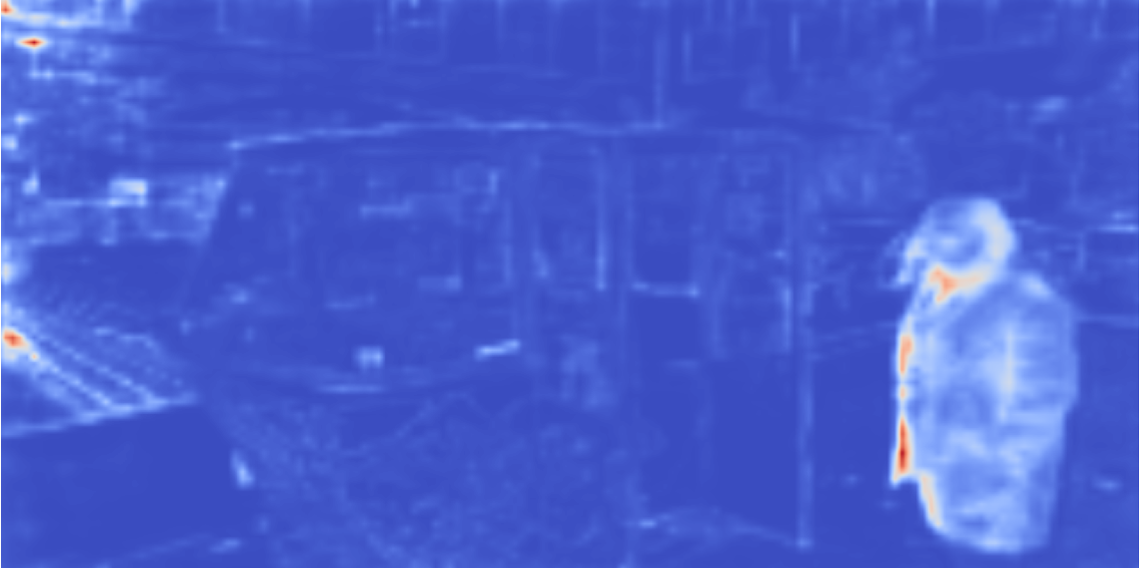}
    \caption{$\beta^t$ (only)}
    \label{fig:3.2}
    \end{subfigure} &
    \begin{subfigure}{0.295\columnwidth}
    \includegraphics[width=\linewidth,height=0.6667\linewidth]{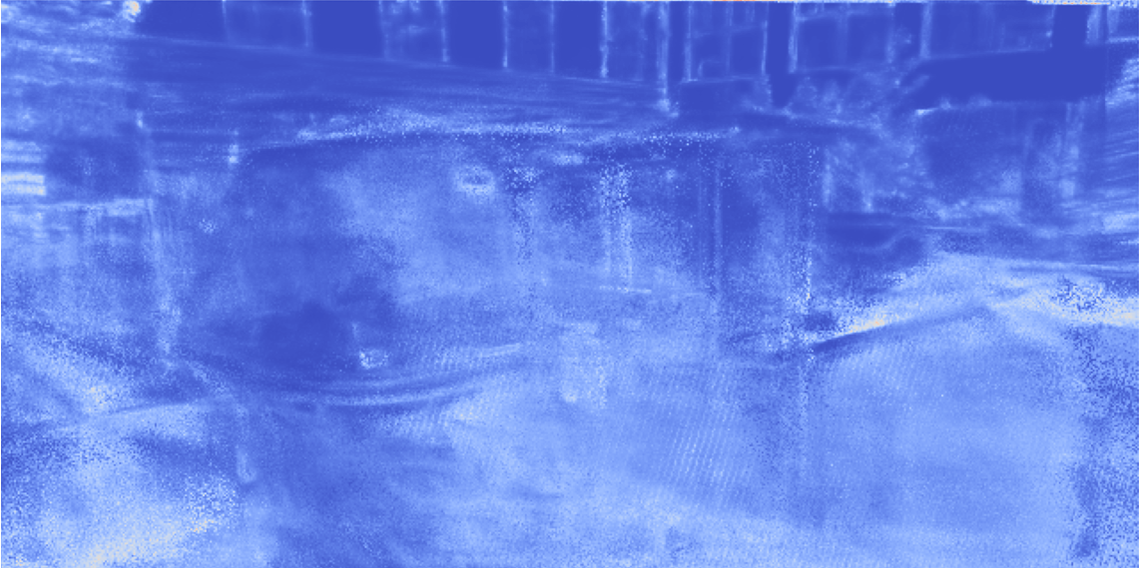}
    \caption{$\beta^s$ (only)}
    \label{fig:3.3}
    \end{subfigure} &
    
    \multirow{2}{*}[2.2cm]{\begin{subfigure}{0.06\columnwidth}
    \includegraphics[width=1.5\linewidth, height=9.5\linewidth]{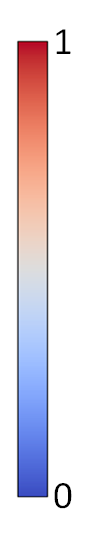}
    \label{fig:3.4}
    \end{subfigure}} 
    \\

    \begin{subfigure}{0.295\columnwidth}
    \includegraphics[width=\linewidth,height=0.6667\linewidth]{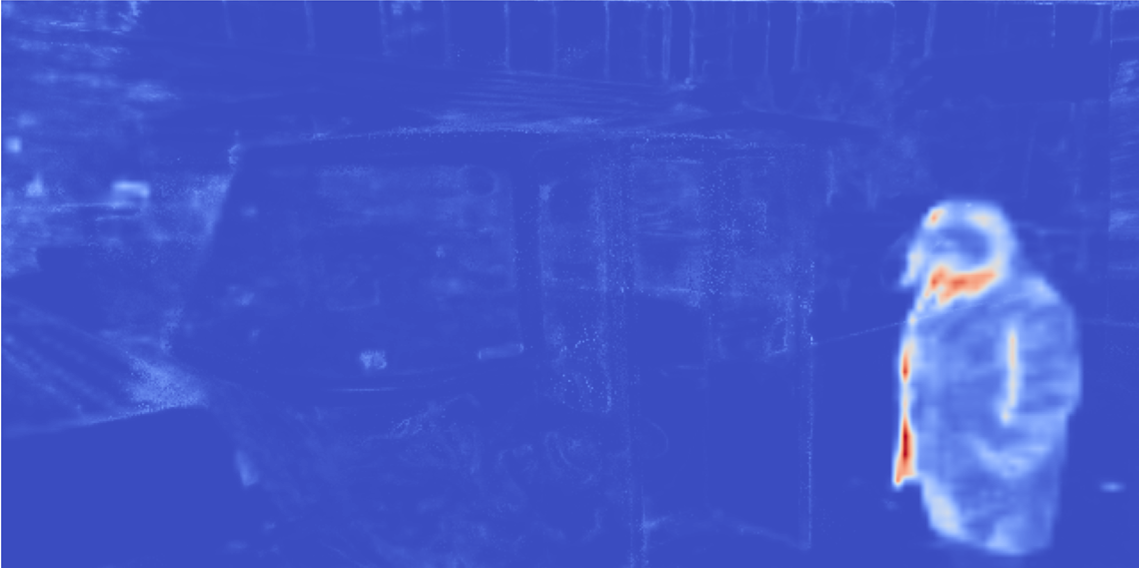}
    \caption{$\beta_{ts}$ }
    \label{fig:3.5}
    \end{subfigure} &
    \begin{subfigure}{0.295\columnwidth}
    \includegraphics[width=\linewidth,height=0.6667\linewidth]{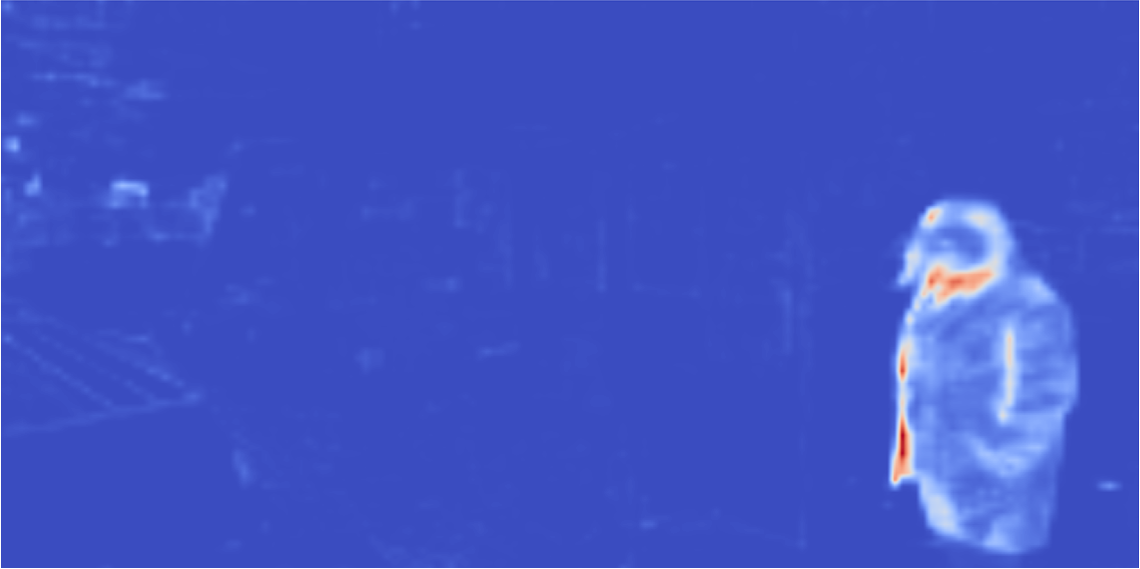}
    \caption{$\beta^t$ (from $\beta_{ts}$) }
    \label{fig:3.6}
    \end{subfigure} &
    \begin{subfigure}{0.295\columnwidth}
    \includegraphics[width=\linewidth,height=0.6667\linewidth]{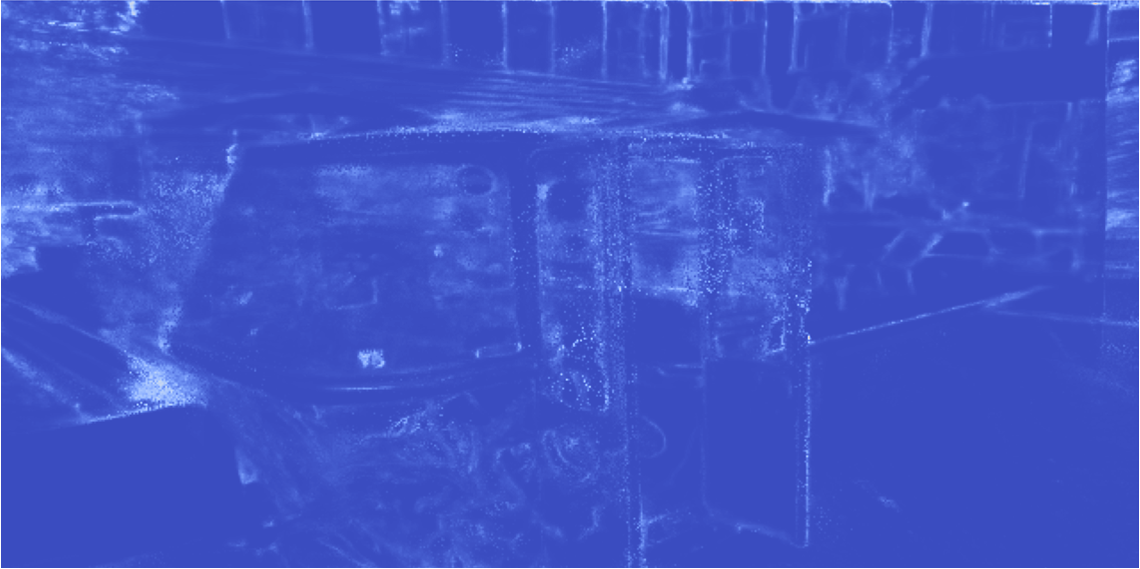}
    \caption{$\beta^s$ (from $\beta_{ts}$)}
    \label{fig:3.7}
    \end{subfigure} & \\
\end{tabular}
\vspace{-0.5em}
\setlength{\belowcaptionskip}{-10pt} 
\caption{\textbf{Visualization results under different uncertainty modeling strategies.} 
(b) Modeling only Target-view Uncertainty $\beta^t$ tends to misjudge static regions as distractors. 
(c) Modeling only Source-view Uncertainty $\beta^s$ fails to localize distractors in the target view.
(d) Jointly modeling both $\beta^s$ and $\beta^t$ and weighted fused into $\beta_{ts}$ under the multi-view uncertainty framework. (e) and (f) are the separate visualizations of these two components to demonstrate their complementary roles: under synergy, $\beta^t$ accurately locates transient distractors, while $\beta^s$  effectively captures static structural discrepancies. Ultimately, $\beta_{ts}$ integrates the advantages of $\beta^t$ and $\beta^s$, enabling more accurate supervision modulation.
The vertical bars show normalized uncertainty (0 to 1), with warmer colors representing higher uncertainty.}

\label{fig:uncertainty}
\end{figure}

\vspace{-1pt}
\subsection{Source-view Uncertainty Prediction}
Due to occlusion and dynamics, the same spatial region may be inconsistently observed different across source views. 
To capture these inconsistencies, we model Source-view Uncertainty in a feed-forward process by leveraging geometric and appearance cues aggregated across source views.

\vspace{0.8pt}
We build the feed-forward network $\mathcal{G}_\theta$ using Transformers from VolRecon \cite{ren2023volrecon} and ReTR \cite{liang2023retr}. Specifically, the View-Transformer integrates multi-source view features, and the Render-Transformer provides rendering weights for each sample point, replacing traditional density-based rendering. To better represent source view information, we also input color samples $\{ c_n^s \}_{n=1}^N$ from the source images. The feed-forward network $\mathcal{G}_\theta$ then outputs each sampled point’s color $c_k \in \mathbb{R}^3$ and Source-view Uncertainty $\beta_k^s \in \mathbb{R}$, which is used to modify Eqn. (\ref{eq:eq2}):
\begin{equation}\label{eq:eq4}
c_k, \beta_k^s = \mathcal{G}_\theta \left( f_n^{s}, c_n^{s}, x_k, d_k \right)
\end{equation}

The aforementioned procedure is applied to all sampling points on each ray, yielding a sequence of distributions $\{ c_k, \beta_k^s \}_{k=1}^K$ along the ray. To map sampled point predictions to image pixels, we model each sampled point as a gaussian distribution with mean $\mu_k = c_k$ and covariance $\Sigma_k = \sigma^2_k \cdot \mathbf{I}$, where $\mathbf{I}$ is the $3\times 3$ identity matrix, implying equal and independent variance across the three color channels, and $\sigma^2_k = \beta^s_k$. Treating all $K$ sample points as components of a Gaussian Mixture Model (GMM) \cite{pan2022activenerf,xue2024neural,liao2024uncertainty}, the pixel distribution $Z$ is represented as a weighted mixture of these gaussians components along the ray:
\begin{equation}\label{eq:eq5}
\text{Prob}(Z) = \sum_{k=1}^K \alpha_k \cdot \mathcal{N}(\mu_k, \Sigma_k)
\end{equation}
where $\alpha_k \in [0,1]$ are weights extracted from the Render-Transformer and normalized via a Softmax, satisfying $\sum_{k=1}^K \alpha_k = 1$. Finally, the pixel color $\hat{C}(r)$ and Source-view Uncertainty $\beta^s(r)$ are computed as the weighted mean $\mu_k$ and variance $\sigma_k^2$, respectively:

\vspace{-3pt}
\begin{equation}\label{eq:eq6}
\hat{C}(r) = \sum_{k=1}^K \alpha_k \cdot \mu_k, \quad 
\beta^s(r) = \sum_{k=1}^K \alpha_k^2 \cdot \sigma^2_k
\end{equation}

This GMM formulation provides a unified mechanism for inferring both color and Source-view Uncertainty at the pixel level, enabling uncertainty-aware rendering that reflects the reliability of aggregated multi-view information. The resulting uncertainty map serves as a principled confidence estimate and is directly used for subsequent supervision modulation.

\begin{figure*}[t!]
    \centering
    \setlength{\figurewidth}{0.185\linewidth}
    \begin{tikzpicture}[
        image/.style = {inner sep=0pt, outer sep=1pt, minimum width=\figurewidth, anchor=north west, text width=\figurewidth}, 
        node distance = 1pt and 1pt, every node/.style={font= {\tiny}}, 
        label/.style = {font={\footnotesize\vphantom{p}},anchor=south,inner sep=0pt}, 
        spy using outlines={rectangle, magnification=2, size=0.36\figurewidth}
        ] 

        \node [image] (img-00) {\includegraphics[width=0.85\figurewidth,height=0.5\figurewidth]{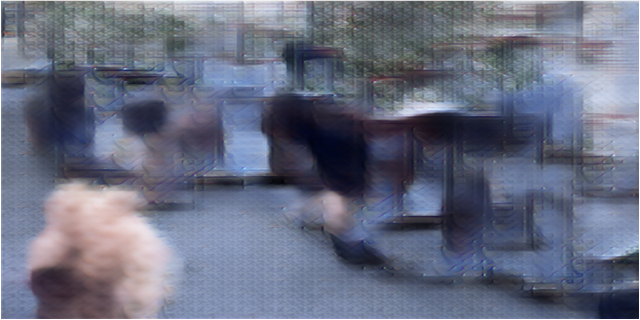}};
        \spy [red,magnification=1.6,size=0.4\figurewidth] on ($(img-00.south) + (-0.365\figurewidth,+0.14\figurewidth)$) in node [below] at ($(img-00.south) + (-0.295\figurewidth,-0.07)$); 
        \node [below=of img-00.south,yshift = -0.6,xshift=0.145\figurewidth ,draw=red, line width=1.5pt,inner sep=0pt] (img-001) {\includegraphics[width=0.39\figurewidth,height=0.39
        \figurewidth]{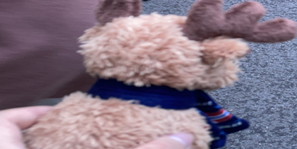}};

        \node [image,right=of img-00,xshift=-0.15\figurewidth] (img-01) {\includegraphics[width=0.85\figurewidth,height=0.5\figurewidth]{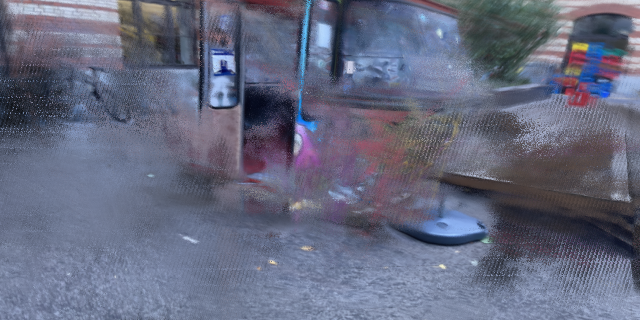}};
        \spy [red,magnification=6,size=0.4\figurewidth] on ($(img-01.north) + (-0.4\figurewidth,-0.07\figurewidth)$) in node [below] at ($(img-01.south) + (-0.295\figurewidth,-0.07)$); 
        \spy [blue,magnification=4,size=0.4\figurewidth] on ($(img-01.north) + (-0.38\figurewidth,-0.15\figurewidth)$) in node [below] at ($(img-01.south) + (0.145\figurewidth,-0.07)$);

        \node [image,right=of img-01,xshift=-0.15\figurewidth] (img-011) {\includegraphics[width=0.85\figurewidth,height=0.5\figurewidth]{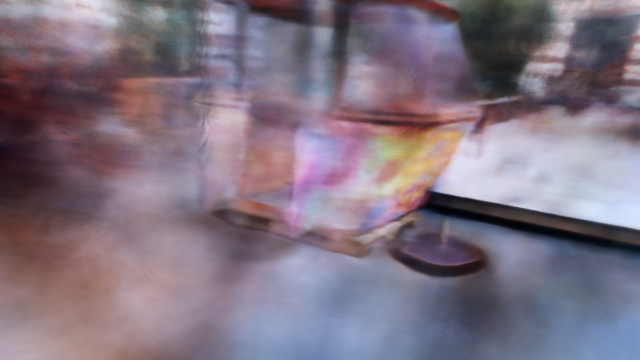}};
        \spy [red,magnification=6,size=0.4\figurewidth] on ($(img-011.north) + (-0.4\figurewidth,-0.07\figurewidth)$) in node [below] at ($(img-011.south) + (-0.295\figurewidth,-0.07)$); 
        \spy [blue,magnification=4,size=0.4\figurewidth] on ($(img-011.north) + (-0.38\figurewidth,-0.15\figurewidth)$) in node [below] at ($(img-011.south) + (0.145\figurewidth,-0.07)$);

        \node [image,right=of img-011,xshift=-0.15\figurewidth] (img-02) {\includegraphics[width=0.85\figurewidth,height=0.5\figurewidth]{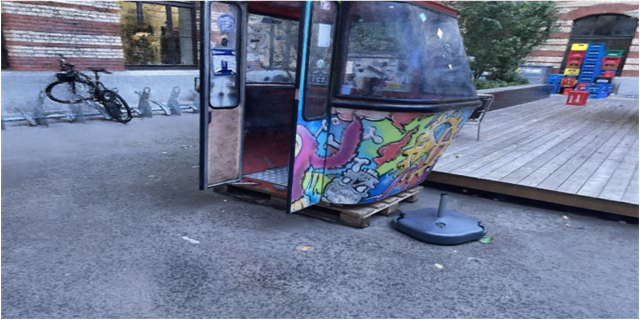}};
        \spy [red,magnification=6,size=0.4\figurewidth] on ($(img-02.north) + (-0.4\figurewidth,-0.07\figurewidth)$) in node [below] at ($(img-02.south) + (-0.295\figurewidth,-0.07)$); 
        \spy [blue,magnification=4,size=0.4\figurewidth] on ($(img-02.north) + (-0.38\figurewidth,-0.15\figurewidth)$) in node [below] at ($(img-02.south) + (0.145\figurewidth,-0.07)$);
        
        \node [image,right=of img-02,xshift=-0.15\figurewidth] (img-03) {\includegraphics[width=0.85\figurewidth,height=0.5\figurewidth]{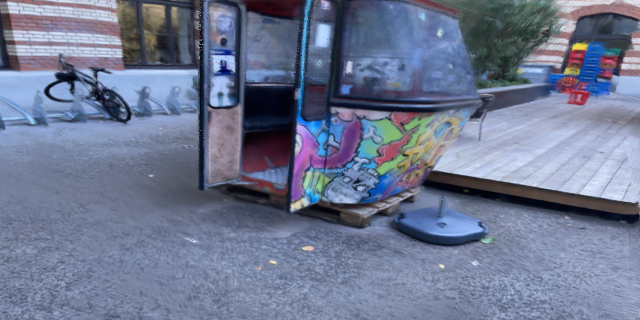}};
         \spy [red,magnification=6,size=0.4\figurewidth] on ($(img-03.north) + (-0.4\figurewidth,-0.07\figurewidth)$) in node [below] at ($(img-03.south) + (-0.295\figurewidth,-0.07)$); 
        \spy [blue,magnification=4,size=0.4\figurewidth] on ($(img-03.north) + (-0.38\figurewidth,-0.15\figurewidth)$) in node [below] at ($(img-03.south) + (0.145\figurewidth,-0.07)$);

        \node [image,right=of img-03,xshift=-0.15\figurewidth] (img-04) {\includegraphics[width=0.85\figurewidth,height=0.5\figurewidth]{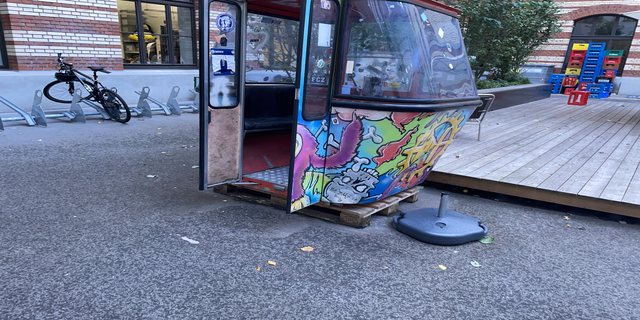}};
        \spy [red,magnification=6,size=0.4\figurewidth] on ($(img-04.north) + (-0.4\figurewidth,-0.07\figurewidth)$) in node [below] at ($(img-04.south) + (-0.295\figurewidth,-0.07)$); 
        \spy [blue,magnification=4,size=0.4\figurewidth] on ($(img-04.north) + (-0.38\figurewidth,-0.15\figurewidth)$) in node [below] at ($(img-04.south) + (0.145\figurewidth,-0.07)$);

        \node [image, below=0.46\figurewidth of img-00.south] (img-10) {\includegraphics[width=0.85\figurewidth,height=0.5\figurewidth]{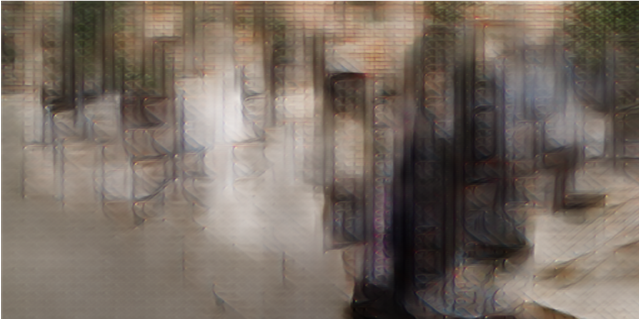}};
        \spy [red,magnification=1.6,size=0.4\figurewidth] on ($(img-10) + (0.09\figurewidth,-0.17)$) in node [below] at ($(img-10.south) + (-0.295\figurewidth,-0.07)$); 
        \node [below=of img-10.south,yshift = -0.6,xshift=0.12\figurewidth ,draw=red, line width=1.5pt,inner sep=0pt] (img-001) {\includegraphics[width=0.39\figurewidth,height=0.39\figurewidth]{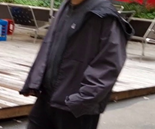}};

        \node [image,right=of img-10,xshift=-0.15\figurewidth] (img-11) {\includegraphics[width=0.85\figurewidth,height=0.5\figurewidth]{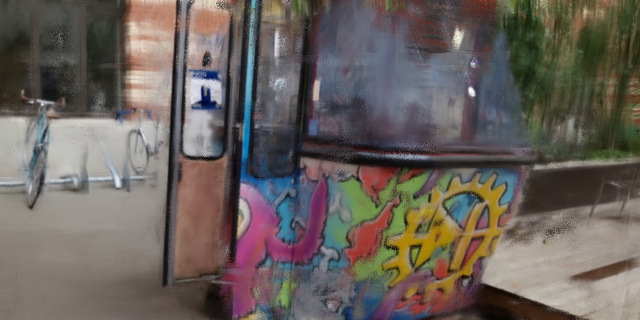}};
        \spy [red,magnification=3,size=0.4\figurewidth] on ($(img-11.north) + (-0.4\figurewidth,-0.12\figurewidth)$) in node [below] at ($(img-11.south) + (-0.295\figurewidth,-0.07)$);  
        \spy [blue,magnification=3,size=0.4\figurewidth] on ($(img-11.east) + (-0.8, - 0.21)$) in node [below] at ($(img-11.south) + (0.145\figurewidth,-0.07)$);

        \node [image,right=of img-11,xshift=-0.15\figurewidth] (img-111) {\includegraphics[width=0.85\figurewidth,height=0.5\figurewidth]{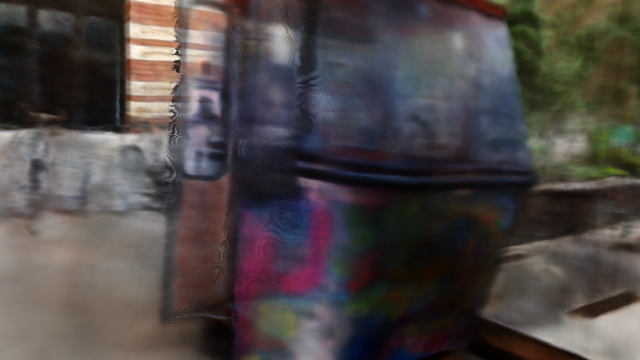}};
        \spy [red,magnification=3,size=0.4\figurewidth] on ($(img-111.north) + (-0.4\figurewidth,-0.12\figurewidth)$) in node [below] at ($(img-111.south) + (-0.295\figurewidth,-0.07)$);  
        \spy [blue,magnification=3,size=0.4\figurewidth] on ($(img-111.east) + (-0.8, - 0.21)$) in node [below] at ($(img-111.south) + (0.145\figurewidth,-0.07)$);

        \node [image,right=of img-111,xshift=-0.15\figurewidth] (img-12) {\includegraphics[width=0.85\figurewidth,height=0.5\figurewidth]{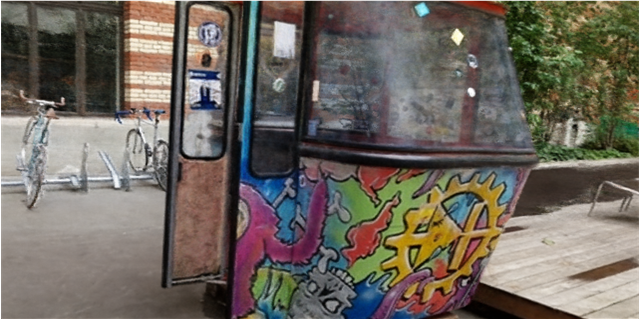}};
         \spy [red,magnification=3,size=0.4\figurewidth] on ($(img-12.north) + (-0.4\figurewidth,-0.12\figurewidth)$) in node [below] at ($(img-12.south) + (-0.295\figurewidth,-0.07)$);  
        \spy [blue,magnification=3,size=0.4\figurewidth] on ($(img-12.east) + (-0.8, - 0.21)$) in node [below] at ($(img-12.south) + (0.145\figurewidth,-0.07)$);
        
        \node [image,right=of img-12,xshift=-0.15\figurewidth] (img-13) {\includegraphics[width=0.85\figurewidth,height=0.5\figurewidth]{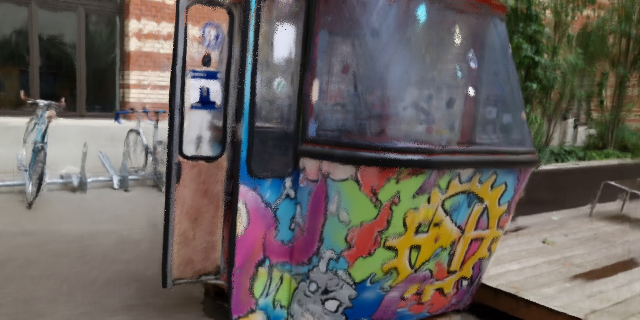}};
         \spy [red,magnification=3,size=0.4\figurewidth] on ($(img-13.north) + (-0.4\figurewidth,-0.12\figurewidth)$) in node [below] at ($(img-13.south) + (-0.295\figurewidth,-0.07)$);  
        \spy [blue,magnification=3,size=0.4\figurewidth] on ($(img-13.east) + (-0.8, - 0.21)$) in node [below] at ($(img-13.south) + (0.145\figurewidth,-0.07)$);

        \node [image,right=of img-13,xshift=-0.15\figurewidth] (img-14) {\includegraphics[width=0.85\figurewidth,height=0.5\figurewidth]{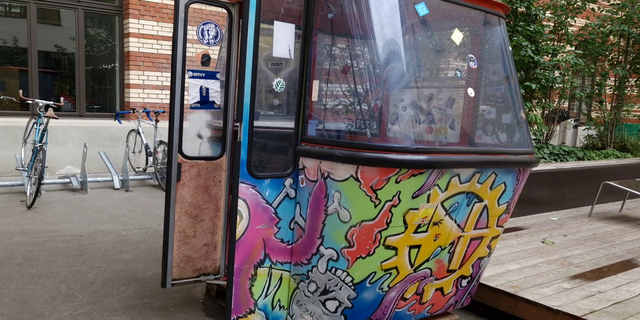}};
        \spy [red,magnification=3,size=0.4\figurewidth] on ($(img-14.north) + (-0.4\figurewidth,-0.12\figurewidth)$) in node [below] at ($(img-14.south) + (-0.295\figurewidth,-0.07)$);  
        \spy [blue,magnification=3,size=0.4\figurewidth] on ($(img-14.east) + (-0.8, - 0.21)$) in node [below] at ($(img-14.south) + (0.145\figurewidth,-0.07)$);
        
        \node[label] (label1) at (img-00.north) {MuRF\textsuperscript{$\dagger$}};
        \node[label] (label2) at (img-01.north) {ReTR};
        \node[label] (label21) at (img-011.north) {UP-NeRF};
        \node[label] (label3) at (img-02.north) {NeRF on-the-go};
        \node[label] (label4) at (img-03.north) {\textbf{Ours}};
        \node[label] (label5) at (img-04.north) {GT Iamge};

        \node[label,rotate=90] (scene1) at ($(img-00.west) + (0,-0.26\figurewidth)$) { Patio-High};
        \node[label,rotate=90] (scene2) at ($(img-10.west) + (0,-0.26\figurewidth)$) { Patio};
    \end{tikzpicture}
    \caption{\textbf{Qualitative comparison on On-the-go after fine-tuning.} MuRF\textsuperscript{$\dagger$} implicitly models inter-view consistency, making it susceptible to transient distractors and causing its rendered results to exhibit distractor artifacts.
    }
    \label{fig:On-the-go}
\end{figure*}
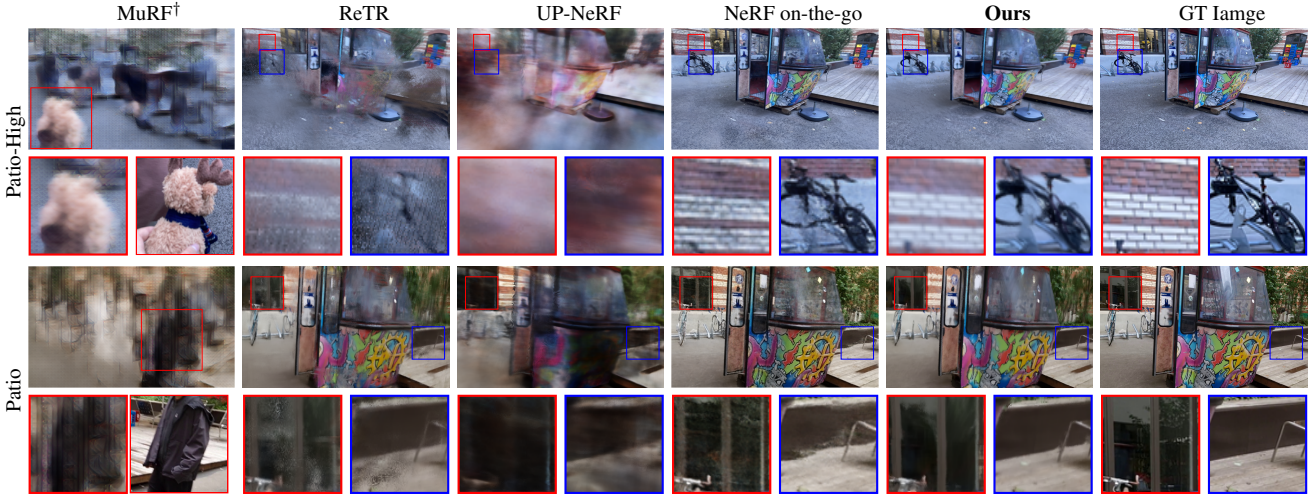


\begin{table*}[t]
\small
\renewcommand{\arraystretch}{0.99}
\tabcolsep 0.022cm
\begin{tabular}{l|c|ccc|ccc|ccc|ccc}
    \toprule
      \multirow{2}{*}{Method}  & 
      \multirow{2}{*}{Setting} & 
      \multicolumn{3}{c|}{Corner} &
      \multicolumn{3}{c|}{Patio} &
      \multicolumn{3}{c|}{Spot} &
      \multicolumn{3}{c}{Patio-High} \\
     & &PSNR↑ & SSIM↑ & LPIPS↓ &PSNR↑ & SSIM↑ & LPIPS↓ &PSNR↑ & SSIM↑ & LPIPS↓ &PSNR↑ & SSIM↑ & LPIPS↓  \\
      
  \midrule
    ReTR (NIPS 2023)& \multirow{3}{*}{\makecell{No optimized \\ per-scene}} &16.33  &0.545 &0.541 &16.39 &0.418 &0.568 &17.43 &0.475 & 0.502 &15.66 &0.327 &0.593   \\

    MuRF (CVPR 2024)& & 13.41 &0.348 &0.692 &11.78 &0.266 & 0.647 &14.18 &0.342 &0.657 &11.88 &0.234 &0.707 \\

    \textbf{MU-GeNeRF (Ours)} & &\textbf{17.96}  &\textbf{0.597}  &\textbf{0.502} &\textbf{18.63} &\textbf{0.543} &\textbf{0.454}  &\textbf{19.35}  &\textbf{0.510} &\textbf{0.469} &\textbf{17.76} &\textbf{0.446}  &\textbf{0.507} \\ \midrule

    ReTR  ft (NIPS 2023) & \multirow{4}{*}{\makecell{Optimized \\ per-scene}} & 19.76 & 0.611 & 0.439 & 17.97 & 0.485& 0.494 & 18.52 & 0.495 & 0.483 & 16.88 & 0.381 & 0.566 \\

    MuRF ft (CVPR 2024) & &13.57 & 0.356 & 0.685 & 12.26 & 0.290 & 0.630 & 15.02 & 0.382 & 0.627 & 12.01 & 0.276 & 0.693 \\

    MuRF\textsuperscript{$\dagger$} ft (CVPR 2024) & &14.03 & 0.363 & 0.677 & 13.22 & 0.286 & 0.616 & 15.15 & 0.384 & 0.621 & 12.42 & 0.280 & 0.688 \\

    UP-NeRF (NIPS 2023) & &19.34 & 0.619 & 0.443 & 15.78 & 0.491 & 0.624 & 16.71 & 0.398 & 0.606 & 14.52 & 0.379 & 0.723 \\

    NeRF on-the-go (CVPR 2024) & &\textbf{23.15} &\textbf{0.751} &\textbf{0.252} & \textbf{21.35} & \textbf{0.717} & \textbf{0.276} & \textbf{23.03} & \textbf{0.727} & \textbf{0.246} & \textbf{20.99} & \textbf{0.687} & \textbf{0.263} \\

    \textbf{MU-GeNeRF (Ours) ft} & & \underline{21.77} & \underline{0.669} & \underline{0.338} & \underline{20.72} & \underline{0.644} & \underline{0.290} & \underline{21.38} & \underline{0.553} & \underline{0.424} & \underline{20.33} & \underline{0.564} & \underline{0.311}\\
    \bottomrule

\end{tabular}
\setlength{\belowcaptionskip}{-8pt}
\caption{\textbf{Quantitative comparison on On-the-go.} ft denotes per-scene fine-tuning. \textsuperscript{†} indicates that the pretrained weights released by the original work are used; methods without this symbol are trained from scratch. }
\label{Tab1}
\end{table*}

\subsection{Target-view Uncertainty Prediction}
We predict a dense uncertainty map from the target view to infer the distractor spatial distribution. Specifically, we first extract semantic features $F^t \in \mathbb{R}^{h \times w \times c}$ from the target image $I^t\in \mathbb{R}^{H \times W \times 3}$ using pre-trained DINOv2 \cite{oquab2023dinov2}. Then, the $F^t$ is subsequently processed by a decoder $\mathcal{F}_\theta$ composed of a CNN and MLP, to generate a dense Target-view Uncertainty map $\beta^t$:
\begin{equation}\label{eq:eq7}
\beta^t(u,v) = \mathcal{U}(\mathcal{F}_\theta(F^t))(u,v), \quad \beta^t \in \mathbb{R}^{H \times W \times1}
\end{equation}
where $\mathcal{U}(\cdot)$ denotes the upsampling function (e.g., bilinear) that resizes the $\mathcal{F}_\theta$ output from $h \times w$ to $H \times W$.  
During training, we employ an extended patch strategy to sample rays from the target image and retrieve the corresponding value $\beta^t(r)$ from $\beta^t$ to serve as the weighting term for heteroscedastic reconstruction loss.
Since $\beta^t$ is predicted by an independent decoder $\mathcal{F}_\theta$ and is not part of the feed-forward process, it is not involved in inference.

Compared with NeRF on-the-go, which estimates uncertainty per ray through a decoupled training procedure, our approach directly predicts a dense uncertainty map from the target image. This fully leverages the spatial modeling capability of CNNs and naturally supports end-to-end optimization with the GeNeRF backbone, simplifying training and improving robustness.

\subsection{Multi-view Uncertainty-Guided Robust Distractor Suppression}
Although both Source-view and Target-view Uncertainties can individually modulate supervision, using either in isolation presents inherent limitations. As shown in Fig. \ref{fig:uncertainty}, the Source-view Uncertainty, derived from the feed-forward process, cannot localize transient distractors in the target view; 
whereas the Target-view Uncertainty, unable to distinguish the sources of reconstruction error, tends to misjudge inconsistent static structures.

Accordingly, we integrate both uncertainties into a heteroscedastic reconstruction loss. Since their origins are different and possess complementary characteristics, they form an implicit synergy under the heteroscedastic framework, effectively compensating for each component’s failure modes when modeled alone, achieving more robust distractor suppression and geometric modeling.
Consequently, the original Eq. \ref{eq:eq3} is reformulated as:

\begin{equation}\label{eq:eq8}
\begin{split}
\mathcal{L}_{\text{Multi-uncer}} = & \frac{\mathcal{L}_{\text{SSIM}}(P(r), \hat{P}(r)) + \mathcal{L}_{\text{MSE}}(P(r), \hat{P}(r))}{2\beta_{ts}^2(r)} \\
& + \lambda \log \beta_{ts}(r)
\end{split}
\end{equation}

where $\beta_{ts} = \omega \cdot \beta^t + (1-\omega) \cdot \beta^s$, $\omega$ is weight coefficient. $P(r)$ and $\hat{P}(r)$ are patches from the ground truth $C(r)$ and rendered images $\hat{C}(r)$, respectively. The loss comprises both Structural Similarity Index Measure (SSIM) \cite{wang2004image} and Mean Squared Error (MSE) terms.

Notably, patch-based SSIM loss is indispensable in our Multi-view Uncertainty framework: 
Both uncertainty components, $\beta^s$ and $\beta^t$, rely on capturing locally correlated structural or semantic variations, which require spatially consistent supervision to be effectively learned.
$\beta^s$ measures the structural conflicts in geometry and appearance across source views (e.g., occlusion boundaries), while $\beta^t$ perceives semantic anomalies within the target view (e.g., transient distractors), both of which exhibit strong spatial correlations. 
However, the pixel-wise MSE loss lacks structural and spatial context awareness, meaning its gradients are easily dominated by isolated noise.
In contrast, the patch-based SSIM loss provides spatially smooth and context-aware gradients, which encourages the model to learn coherent uncertainty distributions, prevents overfitting to isolated noise, and thereby substantially enhances the reliability of estimations and the robustness of the overall training process.

\begin{figure}[t!]
    \centering
    \setlength{\figurewidth}{0.23\linewidth}
    \begin{tikzpicture}[
        image/.style = {inner sep=0pt, outer sep=1pt, minimum width=\figurewidth, anchor=north west, text width=\figurewidth}, 
        node distance = 1pt and 1pt, every node/.style={font= {\tiny}}, 
        label/.style = {font={\footnotesize\vphantom{p}},anchor=south,inner sep=0pt}, 
        spy using outlines={rectangle, magnification=2, size=0.36\figurewidth}
        ] 

        \node [image] (img-00) {\includegraphics[width=\figurewidth,height=0.6667\figurewidth]{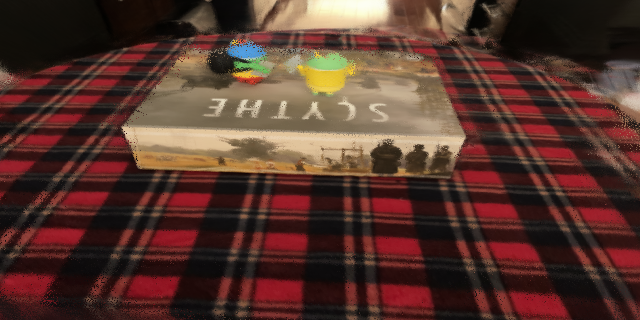}};
        \spy [red,magnification=3,size=0.47\figurewidth] on ($(img-00.north) + (0.24\figurewidth,-0.19)$) in node [below] at ($(img-00.south) + (-0.26\figurewidth,-0.07)$); 
        \spy [blue,magnification=3,size=0.47\figurewidth] on ($(img-00.east) + (-0.19, + 0.3)$) in node [below] at ($(img-00.south) + (0.26\figurewidth,-0.07)$);

        \node [image,right=of img-00,xshift=0.003\figurewidth] (img-01) {\includegraphics[width=\figurewidth,height=0.6667\figurewidth]{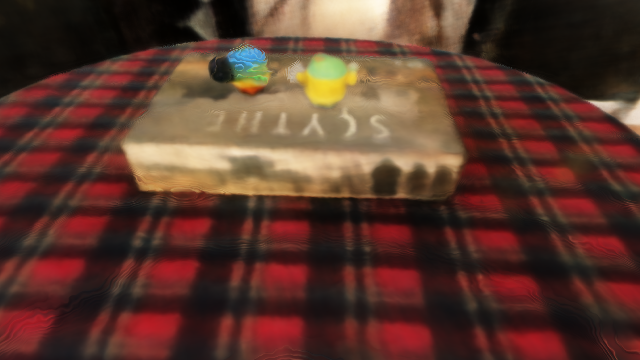}};
         \spy [red,magnification=3,size=0.47\figurewidth] on ($(img-01.north) + (0.24\figurewidth,-0.19)$) in node [below] at ($(img-01.south) + (-0.26\figurewidth,-0.07)$); 
        \spy [blue,magnification=3,size=0.47\figurewidth] on ($(img-01.east) + (-0.19, + 0.3)$) in node [below] at ($(img-01.south) + (0.26\figurewidth,-0.07)$);

        \node [image,right=of img-01,xshift=0.01\figurewidth] (img-02) {\includegraphics[width=\figurewidth,height=0.6667\figurewidth]{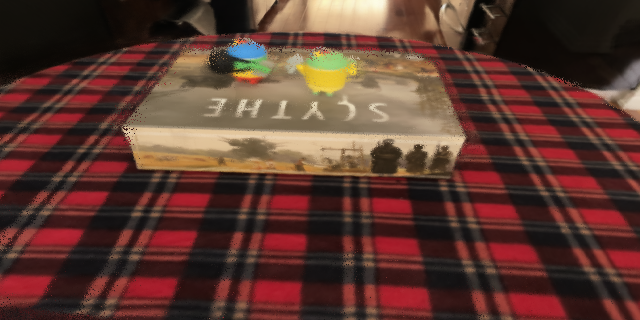}};
        \spy [red,magnification=3,size=0.47\figurewidth] on ($(img-02.north) + (0.24\figurewidth,-0.19)$) in node [below] at ($(img-02.south) + (-0.26\figurewidth,-0.07)$); 
        \spy [blue,magnification=3,size=0.47\figurewidth] on ($(img-02.east) + (-0.19, + 0.3)$) in node [below] at ($(img-02.south) + (0.26\figurewidth,-0.07)$);
        
        \node [image,right=of img-02,xshift=0.01\figurewidth] (img-03) {\includegraphics[width=\figurewidth,height=0.6667\figurewidth]{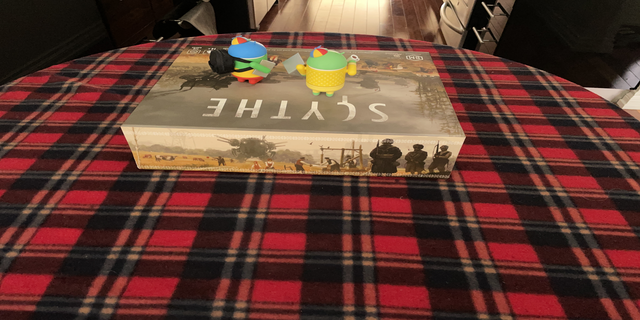}};
        \spy [red,magnification=3,size=0.47\figurewidth] on ($(img-03.north) + (0.24\figurewidth,-0.19)$) in node [below] at ($(img-03.south) + (-0.26\figurewidth,-0.07)$); 
        \spy [blue,magnification=3,size=0.47\figurewidth] on ($(img-03.east) + (-0.19, + 0.3)$) in node [below] at ($(img-03.south) + (0.26\figurewidth,-0.07)$);

        \node [image, below=0.57\figurewidth of img-00.south] (img-10) {\includegraphics[width=\figurewidth,height=0.6667\figurewidth]{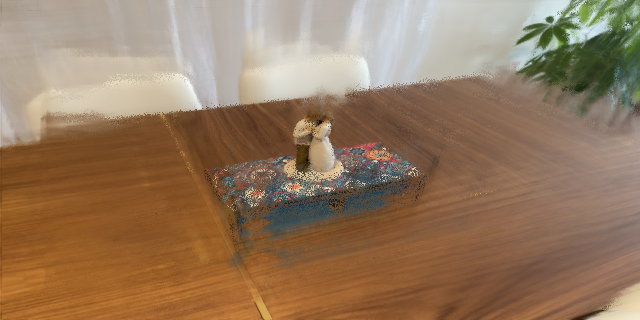}};
        \spy [red,magnification=3,size=0.47\figurewidth] on ($(img-10.west) + (0.3\figurewidth,+0.2)$) in node [below] at ($(img-10.south) + (-0.26\figurewidth,-0.07)$); 
        \spy [blue,magnification=3,size=0.47\figurewidth] on ($(img-10.east) + (-0.3, + 0.4)$) in node [below] at ($(img-10.south) + (0.26\figurewidth,-0.07)$);

        \node [image,right=of img-10,xshift=0.01\figurewidth] (img-11) {\includegraphics[width=\figurewidth,height=0.6667\figurewidth]{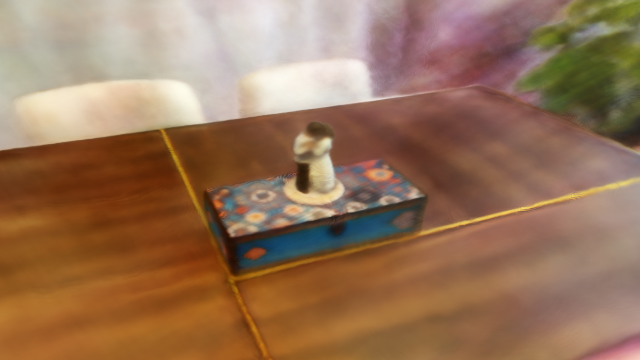}};
        \spy [red,magnification=3,size=0.47\figurewidth] on ($(img-11.west) + (0.3\figurewidth,+0.2)$) in node [below] at ($(img-11.south) + (-0.26\figurewidth,-0.07)$); 
        \spy [blue,magnification=3,size=0.47\figurewidth] on ($(img-11.east) + (-0.3, + 0.4)$) in node [below] at ($(img-11.south) + (0.26\figurewidth,-0.07)$);

        \node [image,right=of img-11,xshift=0.01\figurewidth] (img-12) {\includegraphics[width=\figurewidth,height=0.6667\figurewidth]{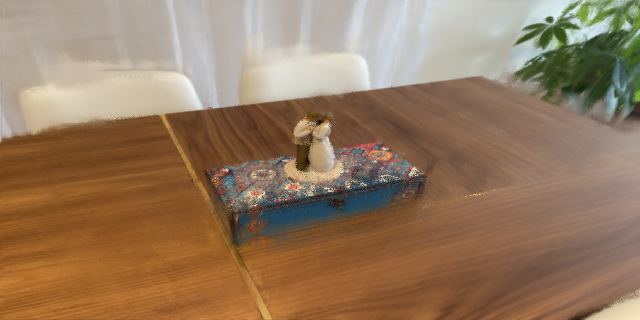}};
        \spy [red,magnification=3,size=0.47\figurewidth] on ($(img-12.west) + (0.3\figurewidth,+0.2)$) in node [below] at ($(img-12.south) + (-0.26\figurewidth,-0.07)$); 
        \spy [blue,magnification=3,size=0.47\figurewidth] on ($(img-12.east) + (-0.3, + 0.4)$) in node [below] at ($(img-12.south) + (0.26\figurewidth,-0.07)$);
        
        \node [image,right=of img-12,xshift=0.01\figurewidth] (img-13) {\includegraphics[width=\figurewidth,height=0.6667\figurewidth]{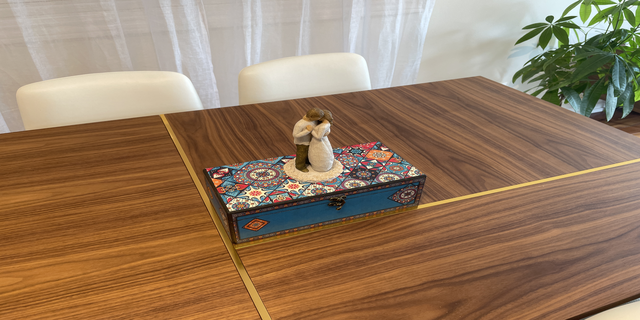}};
        \spy [red,magnification=3,size=0.47\figurewidth] on ($(img-13.west) + (0.3\figurewidth,+0.2)$) in node [below] at ($(img-13.south) + (-0.26\figurewidth,-0.07)$); 
        \spy [blue,magnification=3,size=0.47\figurewidth] on ($(img-13.east) + (-0.3, + 0.4)$) in node [below] at ($(img-13.south) + (0.26\figurewidth,-0.07)$);

        \node[label ] (label1) at (img-00.north) {ReTR};
        \node[label ] (label2) at (img-01.north) {UP-NeRF};
        \node[label ] (label3) at (img-02.north) {\textbf{Ours}};
        \node[label] (label4) at (img-03.north) {GT Image};

        \node[label,rotate=90] (scene1) at ($(img-00.west) + (0,-0.26\figurewidth)$) { Android};
        \node[label,rotate=90] (scene2) at ($(img-10.west) + (0,-0.26\figurewidth)$) { Statue};
    \end{tikzpicture}
    \caption{\textbf{Qualitative comparison on RobustNeRF after fine-tuning.} 
    }
    \label{fig:robustnerf}
\end{figure}
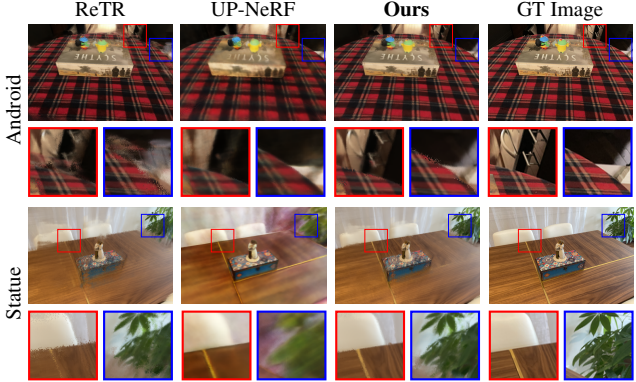

\begin{table}[t]
\renewcommand{\arraystretch}{0.93} 
\tabcolsep 0.023cm 
\begin{tabular}{l|cc|cc}
    \toprule
      \multirow{2}{*}{Method}  & 
      \multicolumn{2}{c|}{Statue} &
      \multicolumn{2}{c}{Android} \\
     & PSNR↑ & SSIM↑ & PSNR↑ & SSIM↑ \\
      
    \midrule
    ReTR (NIPS 2023) &16.84 &0.500  &18.63  &0.531  \\

    MuRF (CVPR 2024) & 12.88 & 0.337 & 11.94 & 0.370 \\

    \textbf{MU-GeNeRF (Ours)} &\textbf{18.28}  &\textbf{0.577}  &\textbf{19.87} &\textbf{0.583}  \\ \midrule
    
    ReTR ft (NIPS 2023) & 18.77 & 0.585 & 20.29 & 0.601 \\

    MuRF ft (CVPR 2024) & 13.10 & 0.351 & 12.34 & 0.389 \\

    MuRF\textsuperscript{$\dagger$} ft (CVPR 2024) & 13.51 & 0.366 & 12.93 & 0.401 \\

    UP-NeRF (NIPS 2023) & 18.10 & 0.629 & 20.45 & 0.664 \\

    NeRF on-the-go (CVPR 2024) & \textbf{21.25} & \textbf{0.732} & \textbf{23.17} & \textbf{0.756} \\

    \textbf{MU-GeNeRF (Ours) ft} & \underline{19.97} & \underline{0.651} & \underline{22.34} & \underline{0.701} \\
    \bottomrule

\end{tabular}
\setlength{\belowcaptionskip}{-10pt}
\caption{\textbf{Quantitative comparison on RobustNeRF.}}
\label{Tab2}
\end{table}

%

\begin{figure}[h]
    \centering
    \setlength{\figurewidth}{0.47\linewidth}
    \begin{tikzpicture}[
        image/.style = {inner sep=0pt, outer sep=1pt, minimum width=\figurewidth, anchor=north west, text width=\figurewidth}, 
        node distance = 1pt and 1pt, every node/.style={font= {\tiny}}, 
        label/.style = {font={\footnotesize\bf\vphantom{p}},anchor=south,inner sep=0pt}, 
        spy using outlines={rectangle, magnification=2, size=0.36\figurewidth}
        ] 

        \node [image] (img-00) {\includegraphics[width=\figurewidth,height=0.5\figurewidth]{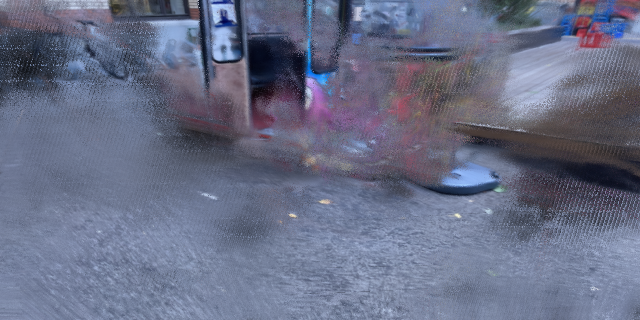}};
        \node[below=6pt of img-00, anchor=center, font=\footnotesize] {w/o Uncertainty Modeling};
        
        \node [image,right=of img-00,xshift=0.003\figurewidth] (img-01) {\includegraphics[width=\figurewidth,height=0.5\figurewidth]{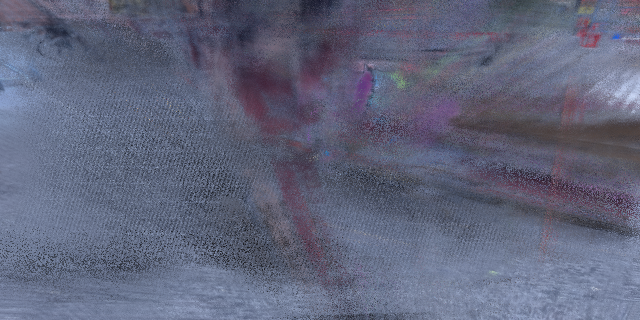}};
        \node[below=6pt of img-01, anchor=center, font=\footnotesize] {w/o SSIM Loss};
        \node [image, below=0.12\figurewidth of img-00.south] (img-10) {\includegraphics[width=\figurewidth,height=0.5\figurewidth]{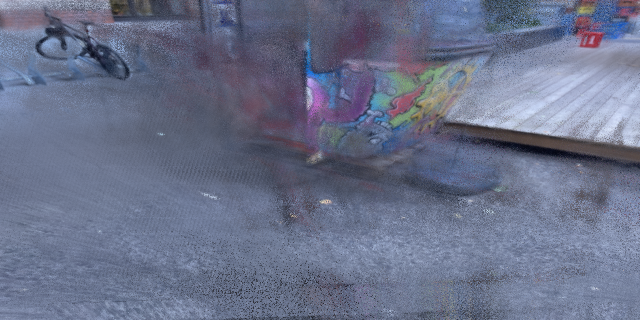}};
        \node[below=6pt of img-10, anchor=center, font=\footnotesize] {w/o Taget-view Uncertainty};
        \node [image,right=of img-10,xshift=0.003\figurewidth] (img-11) {\includegraphics[width=\figurewidth,height=0.5\figurewidth]{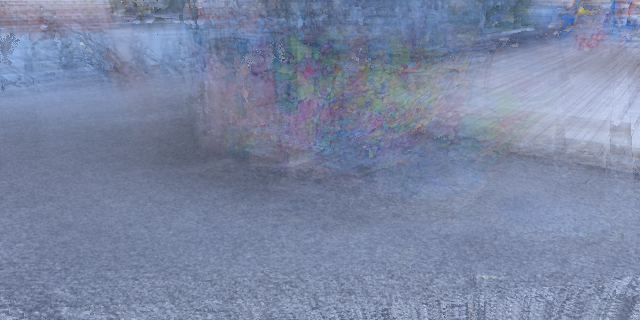}};
        \node[below=6pt of img-11.south, anchor=center, font=\footnotesize] {w/o MSE Loss};
        \node [image, below=0.12\figurewidth of img-10.south] (img-20) {\includegraphics[width=\figurewidth,height=0.5\figurewidth]{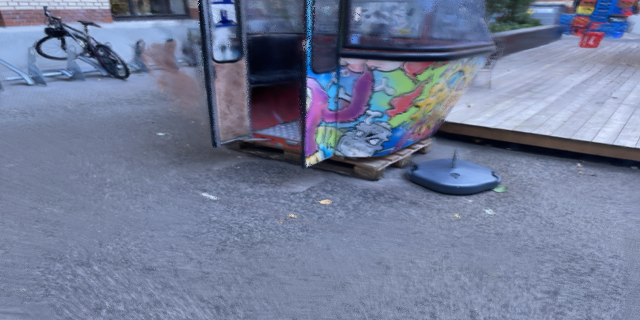}};
         \spy [red,magnification=2.5,size=0.47\figurewidth] on ($(img-20) + (-0.8, + 0.4)$) in node [below] at ($(img-20.south) + (-0.26\figurewidth,-0.07)$); 
        \spy [blue,magnification=3.5,size=0.47\figurewidth] on ($(img-20.north) +(0.016\figurewidth,-0.21\figurewidth)$) in node [below] at ($(img-20.south) + (0.26\figurewidth,-0.07)$);
        \node[below=64pt of img-20.south, anchor=center, font=\footnotesize] {w/o Source-view Uncertainty};
        
        \node [image,right=of img-20,xshift=0.003\figurewidth] (img-21) {\includegraphics[width=\figurewidth,height=0.5\figurewidth]{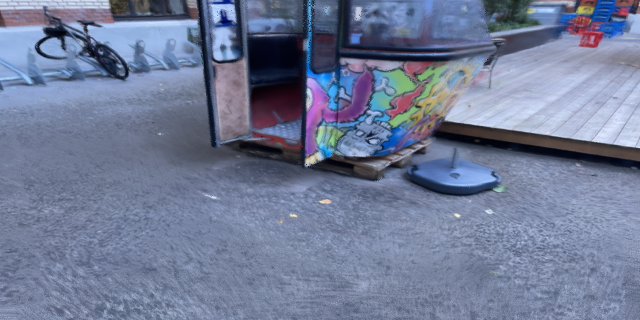}};
         \spy [red,magnification=2.5,size=0.47\figurewidth] on ($(img-21) + (-0.8, + 0.4)$) in node [below] at ($(img-21.south) + (-0.26\figurewidth,-0.07)$); 
        \spy [blue,magnification=3.5,size=0.47\figurewidth] on ($(img-21.north) + (0.016\figurewidth,-0.21\figurewidth)$) in node [below] at ($(img-21.south) + (0.26\figurewidth,-0.07)$);
        \node[below=64pt of img-21.south, anchor=center, font=\footnotesize] {\textbf{Ours (Full)}};

    \end{tikzpicture}
    \setlength{\belowcaptionskip}{-10pt}
    \caption{\textbf{Qualitative comparison of view rendering results under various ablation components.}
    }
    \label{fig:ablation}
\end{figure}
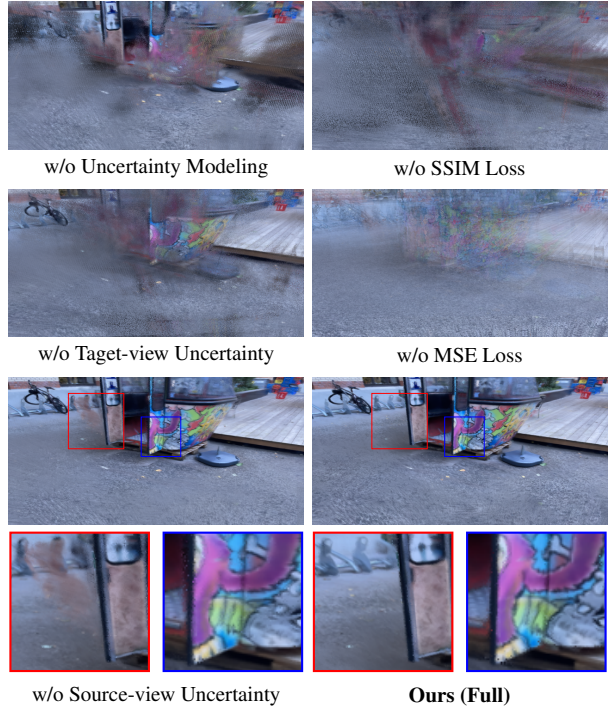



\section{Experiments}	
\subsection{Experimental  Setup}
\vspace{-3pt}
\paragraph{Datasets}
We evaluate our model on two real-world datasets: On-the-go \cite{ren2024nerf} contains various casually captured indoor and outdoor scenes with diverse distractors, while RobustNeRF \cite{sabour2023robustnerf} focuses on indoor scenes with dynamic, non-continuous distractors across frames. We select four scenes from On-the-go and two from RobustNeRF for fine-tuning and reporting, while the remaining scenes are used for generalization training. 

\paragraph{Baselines}
We compare it with several representative methods. ReTR \cite{liang2023retr} and MuRF \cite{xu2024murf} are both GeNeRF-based methods designed for static scenes, without any distractor-aware mechanisms. We further compare with two distractor-free NeRF baselines: UP-NeRF \cite{kim2023up} and NeRF on-the-go \cite{ren2024nerf}. Both methods incorporate uncertainty estimation to remove transient distractors.
Our method and two other GeNeRF methods perform generalization training on multiple scenes, then fine-tune on specific scenes. Two distractor-free NeRF methods train each specific scene from scratch. All experiments are performed on a single NVIDIA A6000 GPU.

\paragraph{Implementation details}
We set $N$ to 4 during training and to 8 during evaluation. 
Among the hyperparameters, $\lambda$ is set to 0.1, and $\omega$ is set to 0.5.
The loss function combines MSE and SSIM losses, with weights of 0.8 and 0.2, respectively.
All images are resized to a uniform resolution of 320×640 to ensure consistency between training and evaluation. 
The batch size is set to 1, with 1024 pixels randomly sampled per batch. Sampling is conducted over $3 \times 3$ patches using a dilation rate of 2 to enhance spatial coverage.
We first conduct 60 epochs ($\sim$2.5 days) of generalization training, then fine-tune for 60K iters on per-scene.

\begin{table}[t]
\renewcommand{\arraystretch}{0.93}
\tabcolsep 0.07cm 
\begin{tabular}{c|c c c c |c c |c c}  
\toprule

\multirow{2}{*}{} & \multicolumn{4}{c|}{} & \multicolumn{2}{c|}{Corner} & \multicolumn{2}{c}{Patio-High} \\  
& $\beta^s$ & $\beta^t$ & $\mathcal{L}_{\text{MSE}}$ & $\mathcal{L}_{\text{SSIM}}$ & PSNR↑ & SSIM↑  & PSNR↑ & SSIM↑  \\

\midrule

0 & & &\checkmark &\checkmark &20.20 & 0.625 & 14.92 & 0.332 \\

1 &\checkmark  & &\checkmark &\checkmark &19.85 &0.618  & 15.33 & 0.341\\

2 & & \checkmark &\checkmark &\checkmark & 20.73 & 0.640 & 19.55 & 0.496 \\ \midrule

3 &\checkmark &\checkmark &\checkmark & &16.84 & 0.563 &13.07 & 0.256\\

4 &\checkmark &\checkmark & & \checkmark &14.87 & 0.497 & 14.10 &0.292\\ \midrule

\textbf{Full} &\checkmark&\checkmark &\checkmark & \checkmark & \textbf{21.77} &  \textbf{0.669} &  \textbf{20.33} &  \textbf{0.564} \\ 
\bottomrule

\end{tabular}
\setlength{\belowcaptionskip}{-8pt}
\caption{\textbf{Quantitative comparison of different ablation components.}}
\label{Tab3}
\end{table}

\begin{figure}[t!]
    \centering
    \setlength{\figurewidth}{0.47\linewidth}
    \begin{tikzpicture}[
        image/.style = {inner sep=0pt, outer sep=1pt, minimum width=\figurewidth, anchor=north west, text width=\figurewidth}, 
        node distance = 1pt and 1pt, every node/.style={font= {\tiny}}, 
        label/.style = {font={\footnotesize\bf\vphantom{p}},anchor=south,inner sep=0pt}, 
        spy using outlines={rectangle, magnification=2, size=0.36\figurewidth}
        ] 
        
        \node [image] (img-00) {\includegraphics[width=\figurewidth,height=0.5\figurewidth]{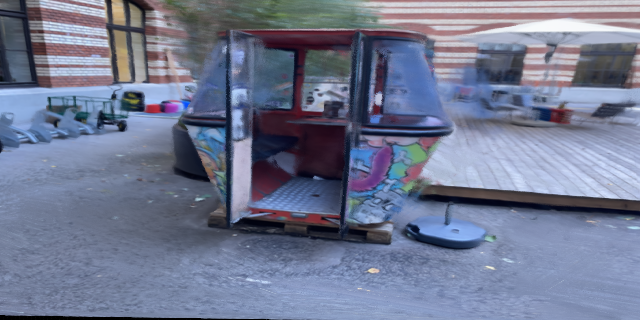}};

        \node[below=0.56\figurewidth of img-00.south, anchor=center, font=\footnotesize] {3-Distractors (PSNR: 19.75)};

                
        \spy [red,magnification=3,size=0.47\figurewidth] on($(img-00) + (-0.13\figurewidth, 0.1)$) in node [below] at ($(img-00.south) + (-0.26\figurewidth,-0.07)$); 
        \spy [blue,magnification=3.6,size=0.47\figurewidth] on ($(img-00.north) + (0.24\figurewidth,-0.15\figurewidth)$) in node [below] at ($(img-00.south) + (0.26\figurewidth,-0.07)$);

        \node [image,right=of img-00,xshift=0.003\figurewidth] (img-01) {\includegraphics[width=\figurewidth,height=0.5\figurewidth]{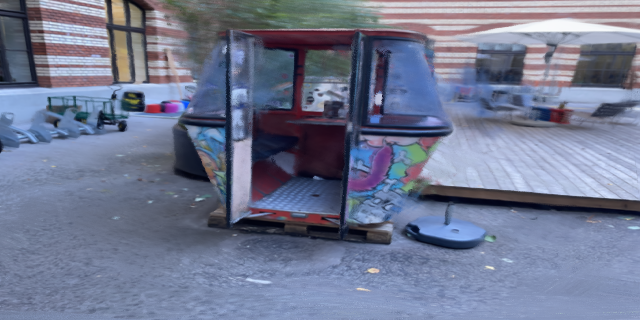}};
        
        
        \node[below=0.56\figurewidth of img-01.south, anchor=center, font=\footnotesize] {2-Distractors (PSNR: 20.43)};
        
        \spy [red,magnification=3,size=0.47\figurewidth] on ($(img-01) + (-0.13\figurewidth, 0.1)$) in node [below] at ($(img-01.south) + (-0.26\figurewidth,-0.07)$); 
        \spy [blue,magnification=3.6,size=0.47\figurewidth] on($(img-01.north) + (0.24\figurewidth,-0.15\figurewidth)$)  in node [below] at ($(img-01.south) + (0.26\figurewidth,-0.07)$);
=
        \node [image,below=0.62\figurewidth of img-00] (img-02) {\includegraphics[width=\figurewidth,height=0.5\figurewidth]{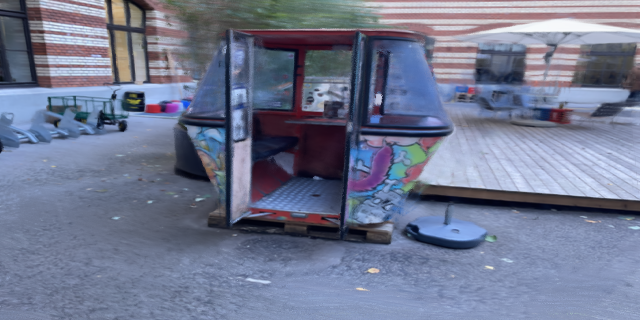}};


        \node[below=0.56\figurewidth of img-02.south, anchor=center, font=\footnotesize] {1-Distractors (PSNR: 21.01)};
        \spy [red,magnification=3,size=0.47\figurewidth] on  ($(img-02) + (-0.13\figurewidth, 0.1)$) in node [below] at ($(img-02.south) + (-0.26\figurewidth,-0.07)$); 
        \spy [blue,magnification=3.6,size=0.47\figurewidth] on ($(img-02.north) + (0.24\figurewidth,-0.15\figurewidth)$)  in node [below] at ($(img-02.south) + (0.26\figurewidth,-0.07)$);
        
        \node [image,right=of img-02,xshift=0.01\figurewidth] (img-03) {\includegraphics[width=\figurewidth,height=0.5\figurewidth]{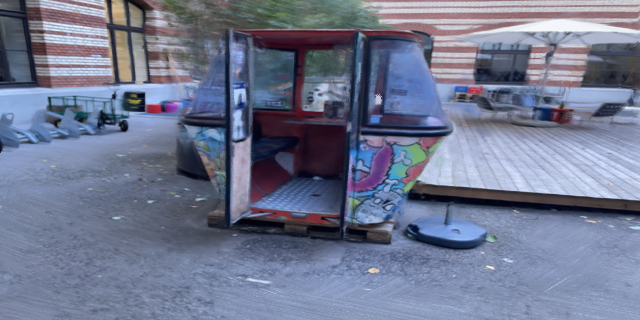}};

        
        \node[below=0.56\figurewidth of img-03.south, anchor=center, font=\footnotesize] {0-Distractors (PSNR: 21.64)};
        \spy [red,magnification=3,size=0.47\figurewidth] on ($(img-03) + (-0.13\figurewidth, 0.1)$) in node [below] at ($(img-03.south) + (-0.26\figurewidth,-0.07)$); 
        \spy [blue,magnification=3.6,size=0.47\figurewidth] on ($(img-03.north) + (0.24\figurewidth,-0.15\figurewidth)$)  in node [below] at ($(img-03.south) + (0.26\figurewidth,-0.07)$);

    \end{tikzpicture}
    \setlength{\belowcaptionskip}{-10pt}
    \caption{\textbf{Rendering results under varying distractor injection levels.} The numeral preceding “-Distractors” denotes the count of source views (out of eight) that include distractors.  
    }
    \label{fig:Robustness}
\end{figure}
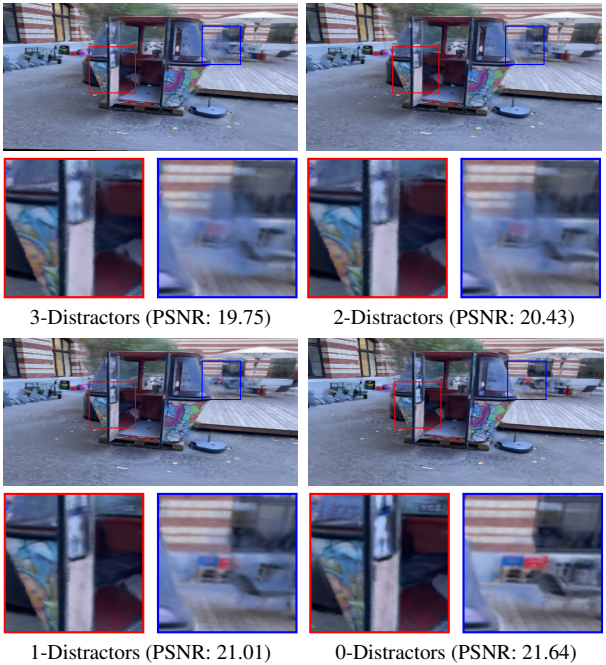

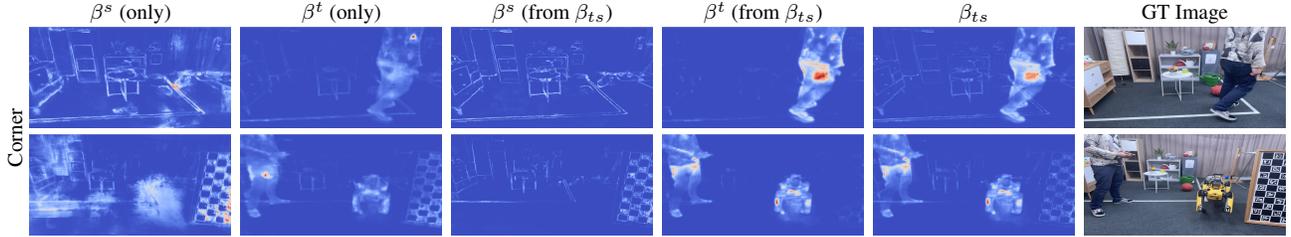
\begin{figure*}[t!]
    \centering
    \setlength{\figurewidth}{0.153\linewidth}
    \begin{tikzpicture}[
        image/.style = {inner sep=0pt, outer sep=1pt, minimum width=\figurewidth, anchor=north west, text width=\figurewidth}, 
        node distance = 1pt and 1pt, every node/.style={font= {\tiny}}, 
        label/.style = {font={\footnotesize\vphantom{p}},anchor=south,inner sep=0pt}, 
        spy using outlines={rectangle, magnification=2, size=0.36\figurewidth}
        ] 

        \node [image] (img-00) {\includegraphics[width=\figurewidth,height=0.5\figurewidth]{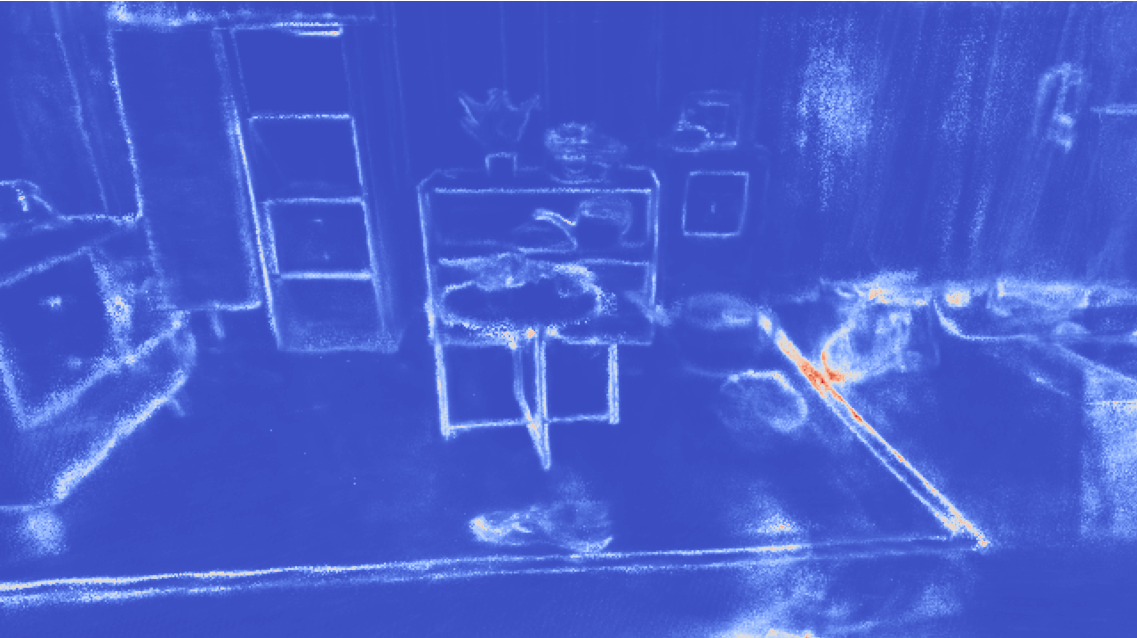}};
        \node [image,right=of img-00,xshift=0.01\figurewidth] (img-01) {\includegraphics[width=\figurewidth,height=0.5\figurewidth]{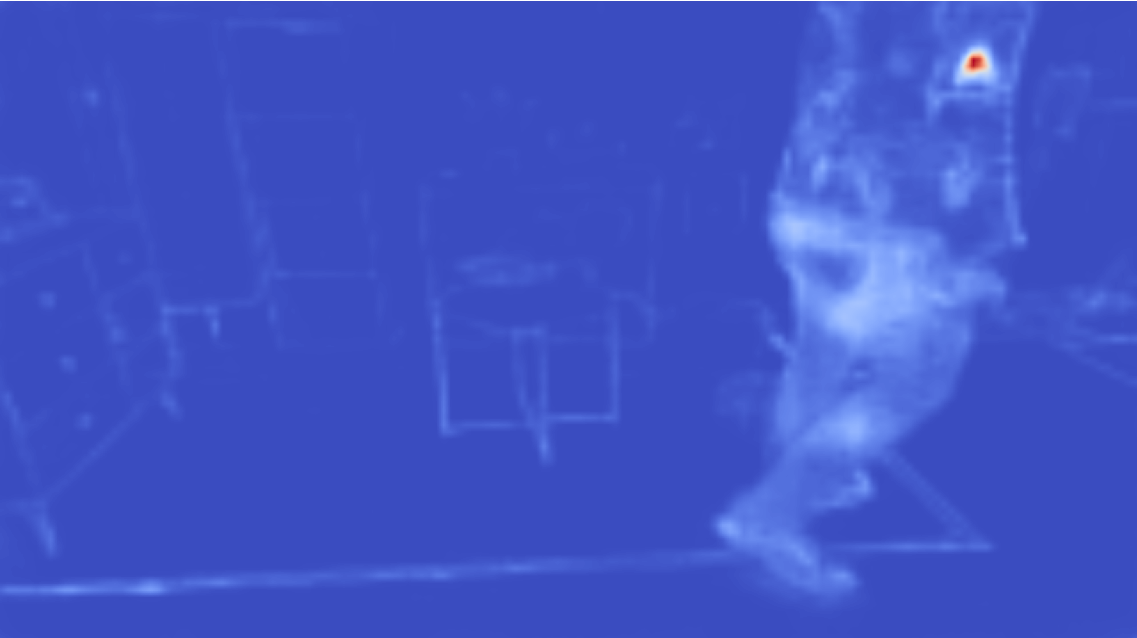}};
        \node [image,right=of img-01,xshift=0.01\figurewidth] (img-02) {\includegraphics[width=\figurewidth,height=0.5\figurewidth]{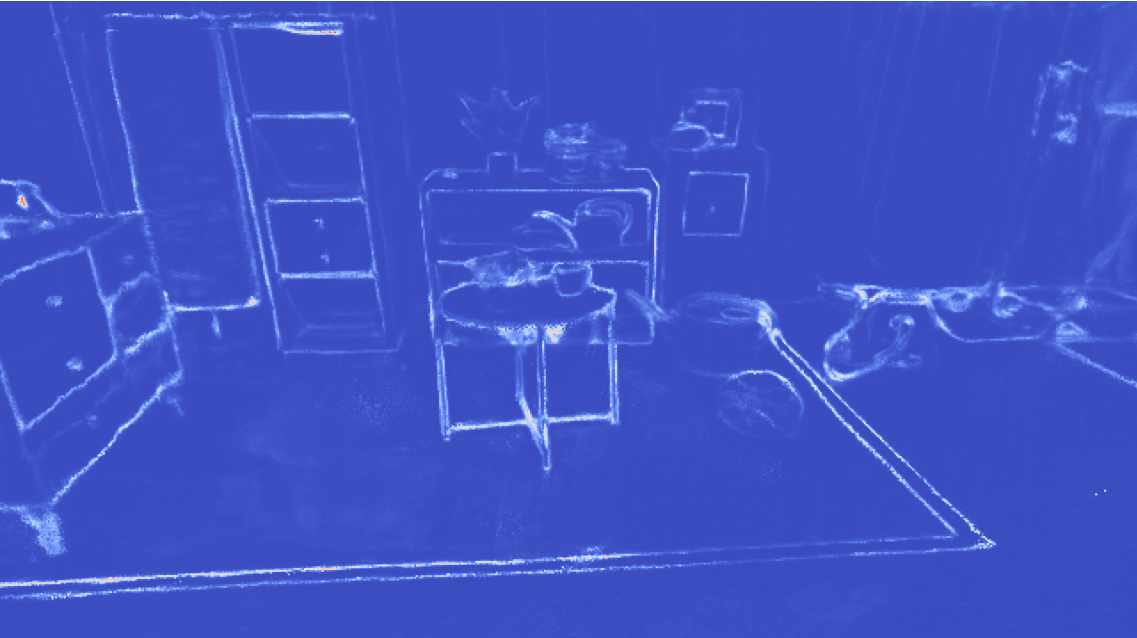}};
        \node [image,right=of img-02,xshift=0.01\figurewidth] (img-03) {\includegraphics[width=\figurewidth,height=0.5\figurewidth]{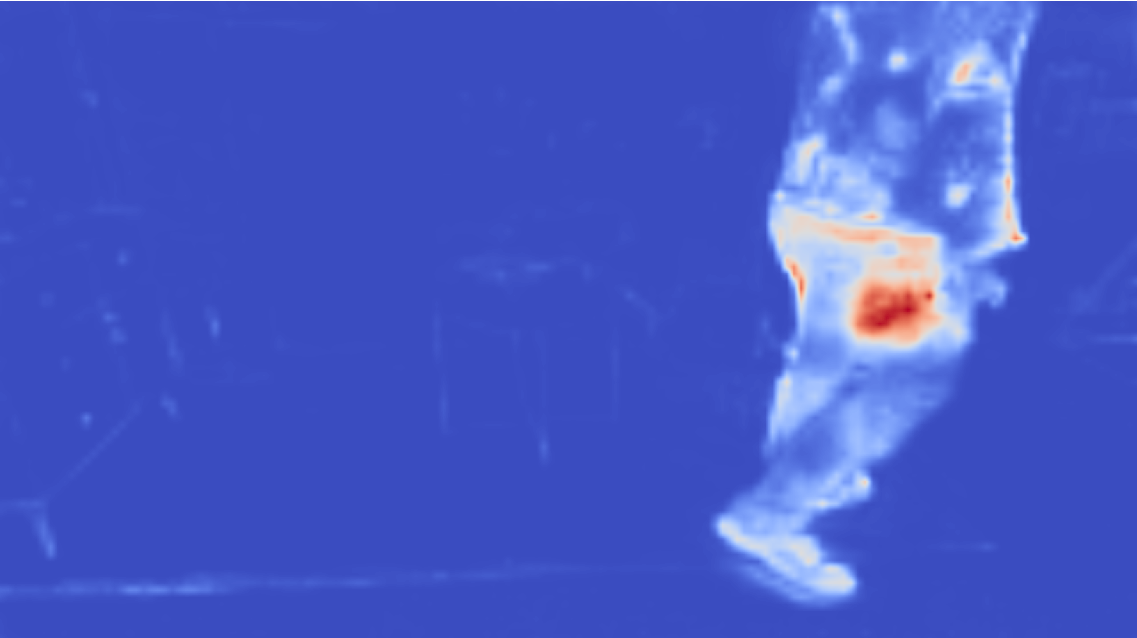}};
        \node [image,right=of img-03,xshift=0.01\figurewidth] (img-04) {\includegraphics[width=\figurewidth,height=0.5\figurewidth]{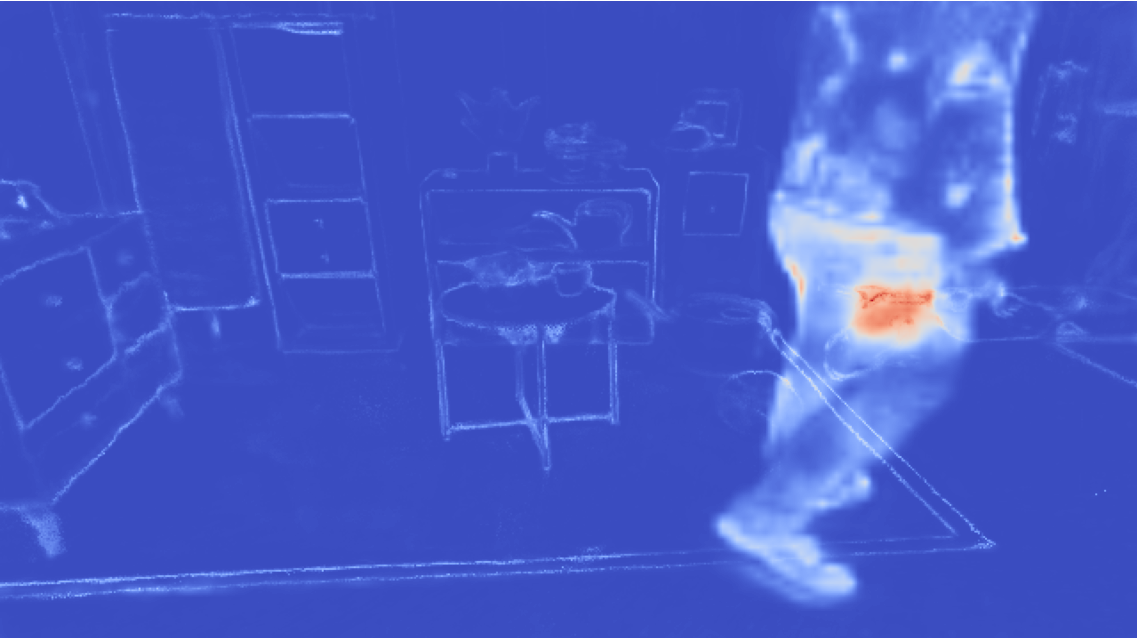}};
        \node [image,right=of img-04,xshift=0.01\figurewidth] (img-05) {\includegraphics[width=\figurewidth,height=0.5\figurewidth]{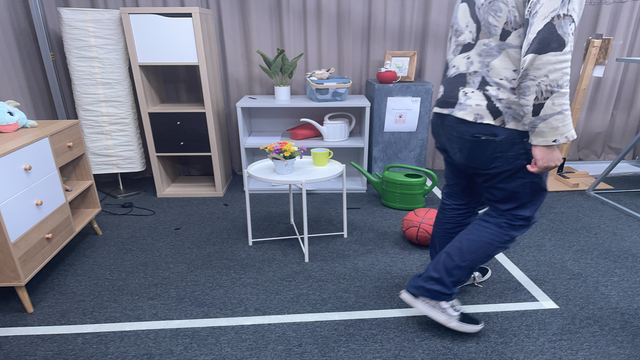}};

        \node [image, below=0.01\figurewidth of img-00.south] (img-10) {\includegraphics[width=\figurewidth,height=0.5\figurewidth]{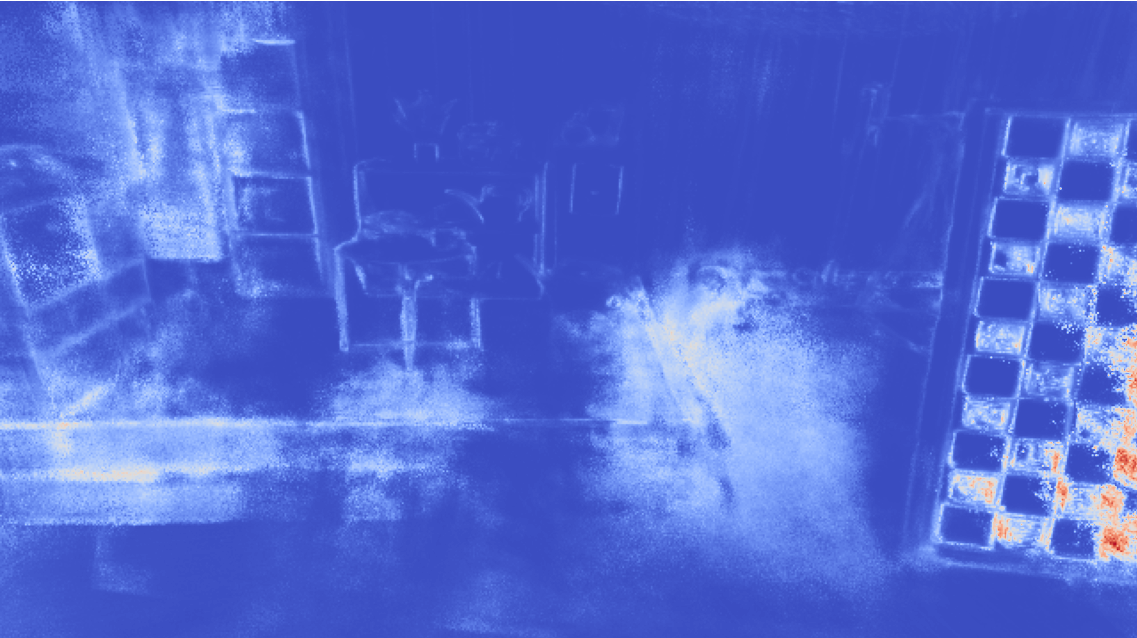}};
        \node [image,right=of img-10,xshift=0.01\figurewidth] (img-11) {\includegraphics[width=\figurewidth,height=0.5\figurewidth]{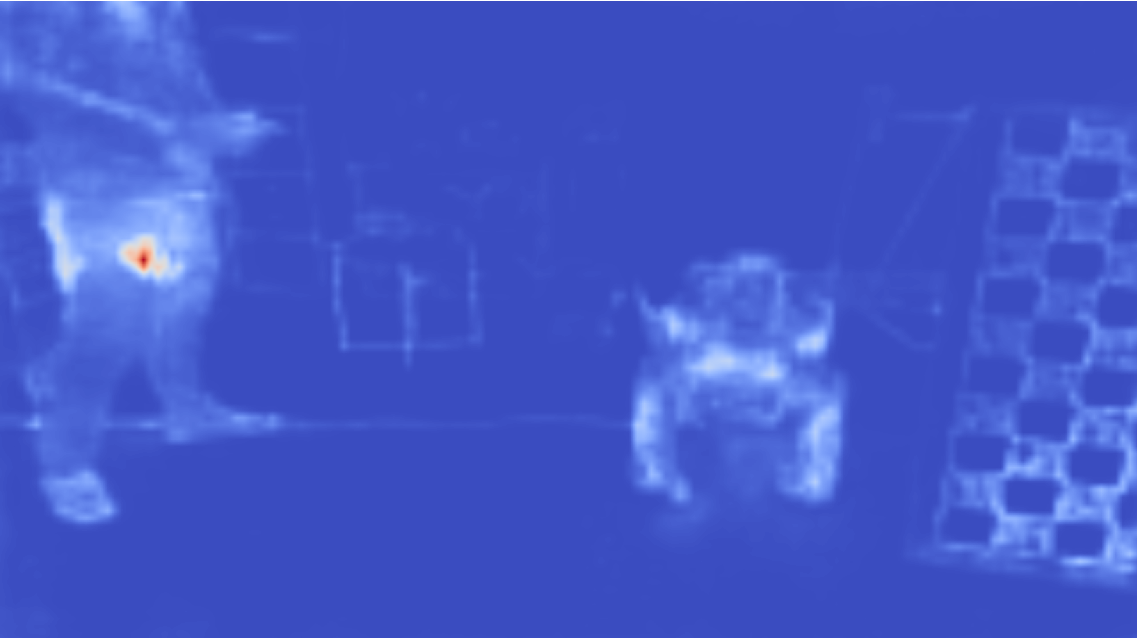}};
        \node [image,right=of img-11,xshift=0.01\figurewidth] (img-12) {\includegraphics[width=\figurewidth,height=0.5\figurewidth]{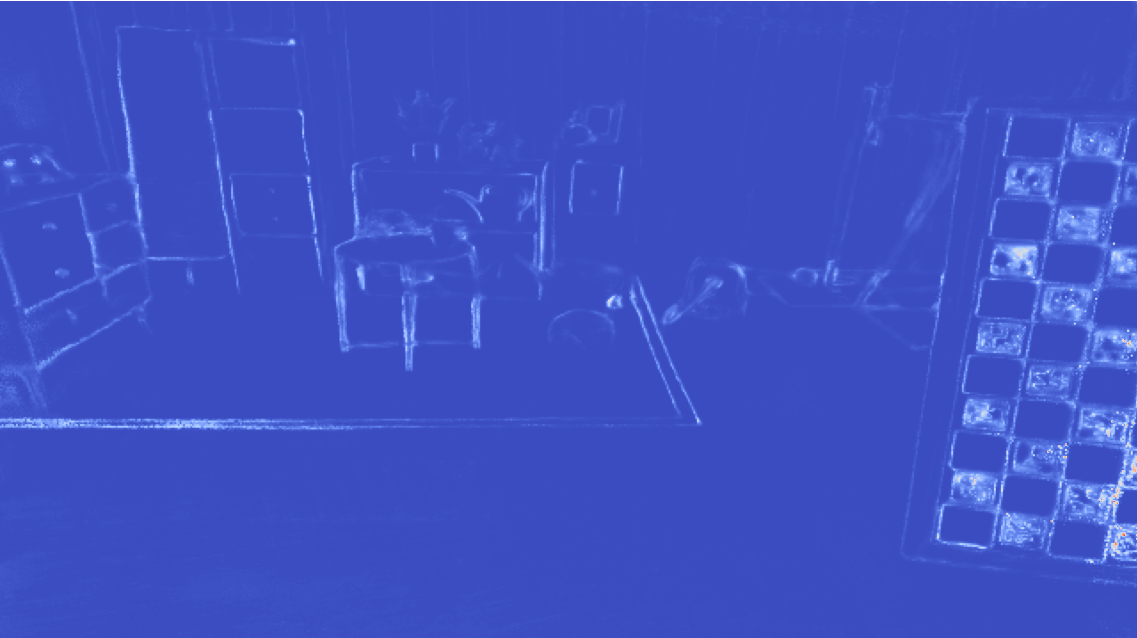}};
        \node [image,right=of img-12,xshift=0.01\figurewidth] (img-13) {\includegraphics[width=\figurewidth,height=0.5\figurewidth]{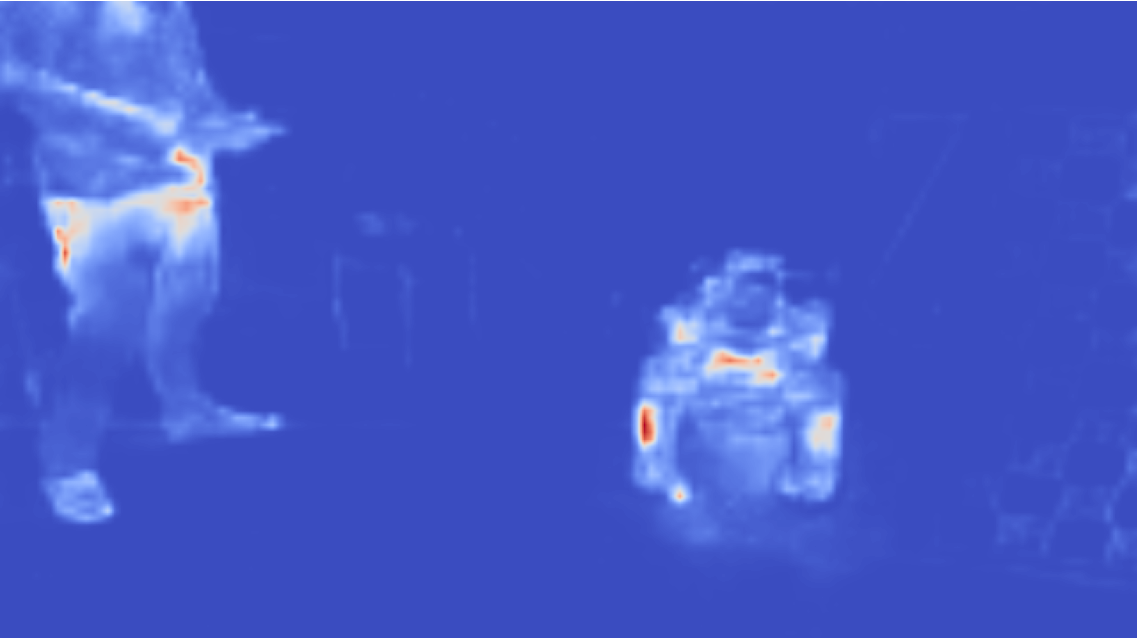}};
        \node [image,right=of img-13,xshift=0.01\figurewidth] (img-14) {\includegraphics[width=\figurewidth,height=0.5\figurewidth]{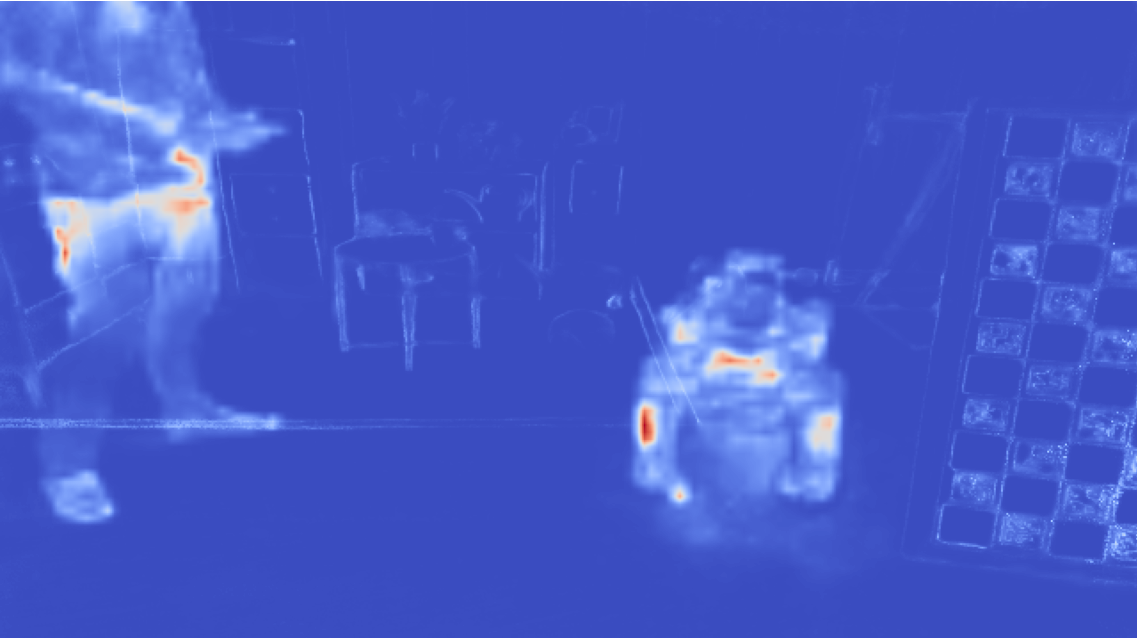}};
        \node [image,right=of img-14,xshift=0.01\figurewidth] (img-15) {\includegraphics[width=\figurewidth,height=0.5\figurewidth]{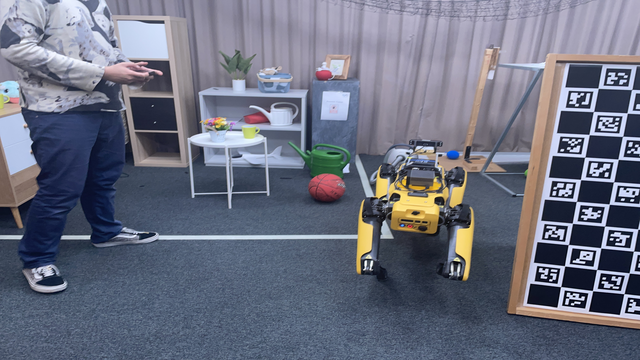}};

        \node[label] (label1) at (img-05.north) {GT Image};
        \node[label] (label2) at (img-00.north) {$\beta^{s}$ (only)};
        \node[label] (label3) at (img-01.north) {$\beta^{t}$ (only)};
        \node[label] (label4) at (img-02.north) {$\beta^s$ (from $\beta_{ts}$)};
        \node[label] (label5) at (img-03.north) {$\beta^t$ (from $\beta_{ts}$)};
        \node[label] (label6) at (img-04.north) {$\beta_{ts}$};
        \node[label,rotate=90] (scene1) at ($(img-00.west) + (0,-0.3\figurewidth)$) { Corner};

    \end{tikzpicture}
    \setlength{\belowcaptionskip}{-5pt}
    \caption{\textbf{Visualization Comparison of Different Uncertainty Modeling Strategies.}}
    \label{fig:Uncertainty-Modeling}
\end{figure*}

\subsection{Quantitative and Qualitative  Evaluation}
Tab. \ref{Tab1} and Tab. \ref{Tab2} present quantitative comparisons with several baselines. 
Before or after per-scene fine-tuning, our method consistently outperforms ReTR and MuRF, demonstrating clear advantages in cross-scene generalization and scene-specific adaptability.
While our method draws on parts of ReTR's architecture, the latter lacks any distractor-aware design and fails to learn robust priors under transient scenes. 
MuRF, designed for large-baseline settings, uses Transformers to implicitly model inter-view consistency. However, it is easily misled by transient distractors, which introduces inconsistent context during view aggregation and significantly degrades reconstruction quality.

Compared to two distractor-free NeRF methods, our method outperforms UP-NeRF, whose performance is limited due to the explicit modeling of transient distractors for separation, which often suffers from inaccurate decoupling in complex scenes.
Although our performance is marginally lower than NeRF on-the-go, the two methods operate under fundamentally different regimes: NeRF on-the-go is a scene-specific NeRF that performs sufficient multi-view optimization within each scene to establish consistency and filter distractors;
In contrast, MU-GeNeRF is a feed-forward GeNeRF that learns transferable aggregation priors via robust supervision, excelling in cross-scene generalization rather than per-scene refinement.
In terms of efficiency, our method requires only about 60K iters ($\sim$2 hours) of fine-tuning to achieve comparable results on some scenes (e.g., Patio and Patio-high), whereas NeRF on-the-go requires 250K iters ($\sim$48 hours) to optimize a per-scene.

Fig. \ref{fig:On-the-go} and Fig. \ref{fig:robustnerf} show qualitative comparisons after the fine-tuning. ReTR fails to preserve structural clarity and detail fidelity, while MuRF incorrectly incorporates distractors, resulting in unsuccessful rendering. In contrast, our method produces more natural visual outputs and achieves image quality on par with NeRF on-the-go, further validating its effectiveness in distractor-aware modeling.

\begin{figure*}[t!]
    \centering
    \setlength{\figurewidth}{0.185\linewidth}
    \begin{tikzpicture}[
        image/.style = {inner sep=0pt, outer sep=1pt, minimum width=\figurewidth, anchor=north west, text width=\figurewidth}, 
        node distance = 1pt and 1pt, every node/.style={font= {\tiny}}, 
        label/.style = {font={\footnotesize\vphantom{p}},anchor=south,inner sep=0pt}, 
        spy using outlines={rectangle, magnification=2, size=0.36\figurewidth}
        ] 

        \node [image] (img-01) {\includegraphics[width=\figurewidth,height=0.5\figurewidth]{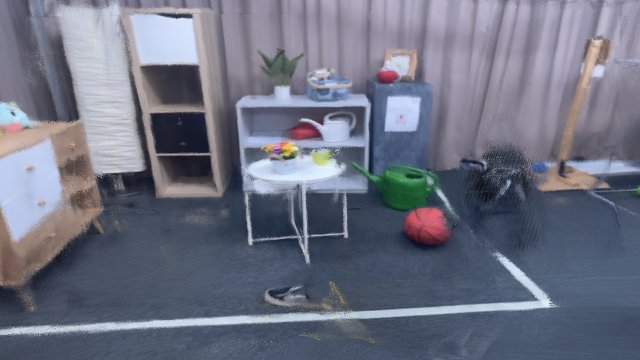}};
        \spy [red,magnification=3,size=0.47\figurewidth] on ($(img-01.west) + (0.12\figurewidth,-0.03\figurewidth)$)  in node [below] at ($(img-01.south) + (-0.26\figurewidth,-0.07)$); 
        \spy [blue,magnification=3,size=0.47\figurewidth] on($(img-01) + (-0.025\figurewidth, -0.16\figurewidth)$)   in node [below] at ($(img-01.south) + (0.26\figurewidth,-0.07)$);
        
        \node [image,right=of img-01,xshift=0.01\figurewidth] (img-02) {\includegraphics[width=\figurewidth,height=0.5\figurewidth]{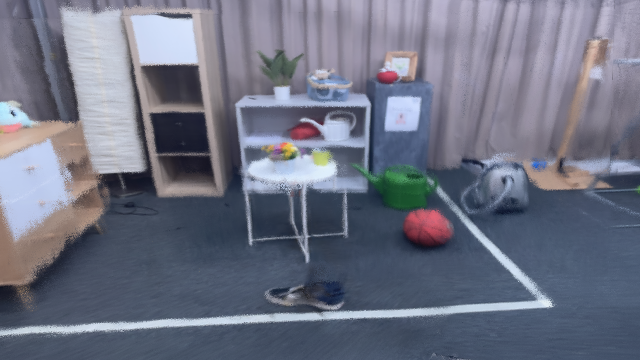}};
        \spy [red,magnification=3,size=0.47\figurewidth] on ($(img-02.west) + (0.12\figurewidth,-0.03\figurewidth)$) in node [below] at ($(img-02.south) + (-0.26\figurewidth,-0.07)$); 
        \spy [blue,magnification=3,size=0.47\figurewidth] on($(img-02) + (-0.025\figurewidth, -0.16\figurewidth)$)   in node [below] at ($(img-02.south) + (0.26\figurewidth,-0.07)$);
        
        \node [image,right=of img-02,xshift=0.01\figurewidth] (img-03) {\includegraphics[width=\figurewidth,height=0.5\figurewidth]{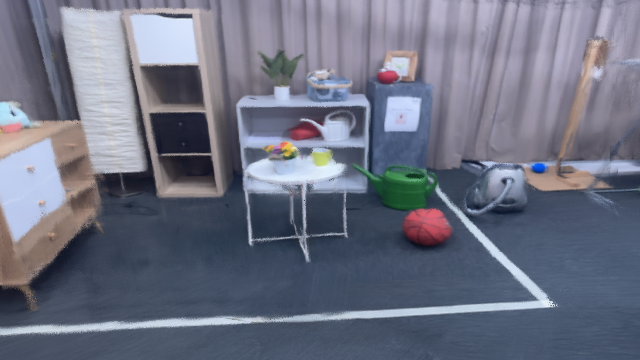}};
        \spy [red,magnification=3,size=0.47\figurewidth] on ($(img-03.west) + (0.12\figurewidth,-0.03\figurewidth)$)  in node [below] at ($(img-03.south) + (-0.26\figurewidth,-0.07)$); 
        \spy [blue,magnification=3,size=0.47\figurewidth] on($(img-03) + (-0.025\figurewidth, -0.16\figurewidth)$)  in node [below] at ($(img-03.south) + (0.26\figurewidth,-0.07)$);
        
        \node [image,right=of img-03,xshift=0.01\figurewidth] (img-04) {\includegraphics[width=\figurewidth,height=0.5\figurewidth]{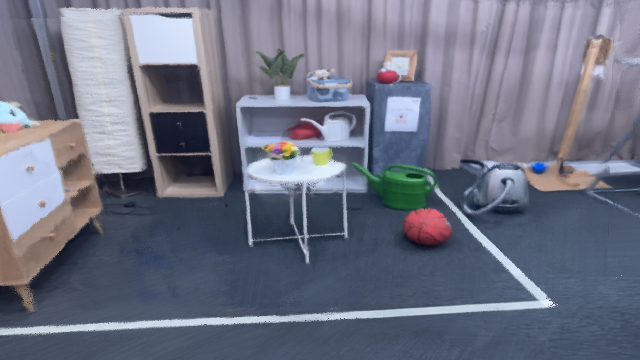}};
        \spy [red,magnification=3,size=0.47\figurewidth] on ($(img-04.west) + (0.12\figurewidth,-0.03\figurewidth)$) in node [below] at ($(img-04.south) + (-0.26\figurewidth,-0.07)$); 
        \spy [blue,magnification=3,size=0.47\figurewidth] on($(img-04) + (-0.025\figurewidth, -0.16\figurewidth)$)  in node [below] at ($(img-04.south) + (0.26\figurewidth,-0.07)$);
        
        \node [image,right=of img-04,xshift=0.01\figurewidth] (img-05) {\includegraphics[width=\figurewidth,height=0.5\figurewidth]{sec/images/fig9/corner/GT-target-34.png}};
        \spy [red,magnification=3,size=0.47\figurewidth] on ($(img-05.west) + (0.12\figurewidth,-0.03\figurewidth)$)  in node [below] at ($(img-05.south) + (-0.26\figurewidth,-0.07)$); 
        \spy [blue,magnification=3,size=0.47\figurewidth] on($(img-05) + (-0.025\figurewidth, -0.16\figurewidth)$)   in node [below] at ($(img-05.south) + (0.26\figurewidth,-0.07)$);

        \node [image, below=0.55\figurewidth of img-01.south] (img-11) {\includegraphics[width=\figurewidth,height=0.5\figurewidth]{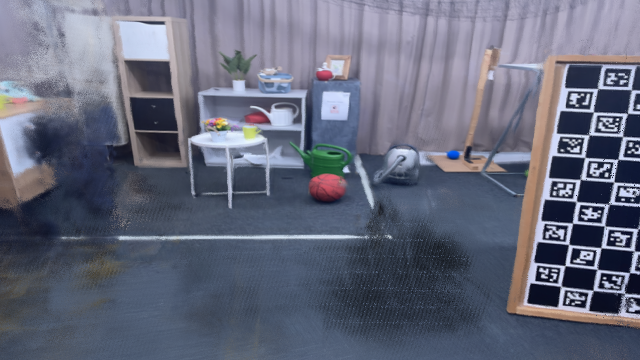}};
        \spy [red,magnification=3,size=0.47\figurewidth] on ($(img-11.west) + (0.1\figurewidth,-0.03\figurewidth)$)  in node [below] at ($(img-11.south) + (-0.26\figurewidth,-0.07)$); 
        \spy [blue,magnification=3,size=0.47\figurewidth] on($(img-11.east) + (-0.1\figurewidth,-0.12\figurewidth)$)  in node [below] at ($(img-11.south) + (0.26\figurewidth,-0.07)$);

        \node [image,right=of img-11,xshift=0.01\figurewidth] (img-12) {\includegraphics[width=\figurewidth,height=0.5\figurewidth]{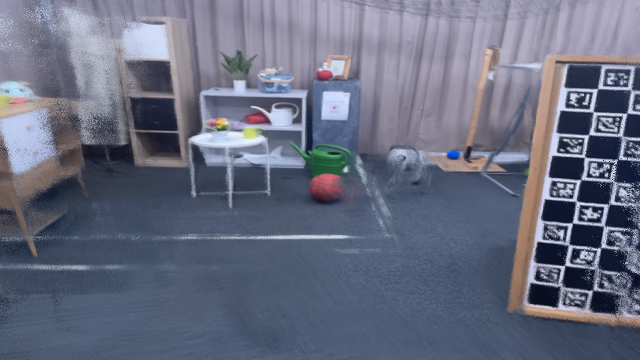}};
        \spy [red,magnification=3,size=0.47\figurewidth] on ($(img-12.west) + (0.1\figurewidth,-0.03\figurewidth)$)  in node [below] at ($(img-12.south) + (-0.26\figurewidth,-0.07)$); 
        \spy [blue,magnification=3,size=0.47\figurewidth] on($(img-12.east) + (-0.1\figurewidth,-0.12\figurewidth)$)  in node [below] at ($(img-12.south) + (0.26\figurewidth,-0.07)$);

        \node [image,right=of img-12,xshift=0.01\figurewidth] (img-13) {\includegraphics[width=\figurewidth,height=0.5\figurewidth]{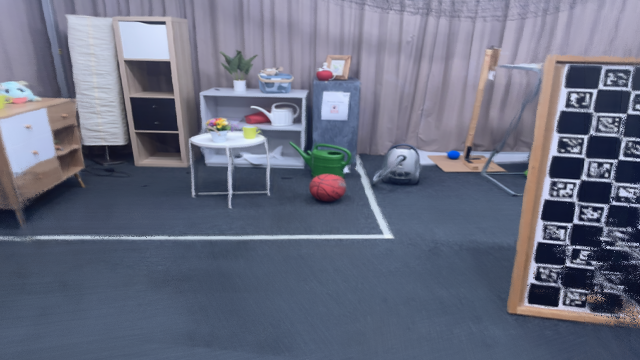}};
        \spy [red,magnification=3,size=0.47\figurewidth] on ($(img-13.west) + (0.1\figurewidth,-0.03\figurewidth)$)  in node [below] at ($(img-13.south) + (-0.26\figurewidth,-0.07)$); 
        \spy [blue,magnification=3,size=0.47\figurewidth] on($(img-13.east) + (-0.1\figurewidth,-0.12\figurewidth)$)  in node [below] at ($(img-13.south) + (0.26\figurewidth,-0.07)$);

        \node [image,right=of img-13,xshift=0.01\figurewidth] (img-14) {\includegraphics[width=\figurewidth,height=0.5\figurewidth]{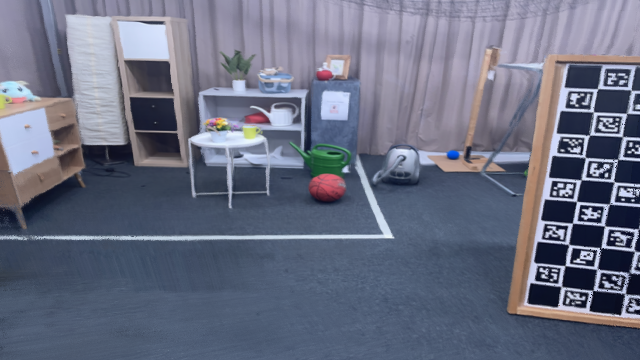}};
        \spy [red,magnification=3,size=0.47\figurewidth] on ($(img-14.west) + (0.1\figurewidth,-0.03\figurewidth)$)  in node [below] at ($(img-14.south) + (-0.26\figurewidth,-0.07)$); 
        \spy [blue,magnification=3,size=0.47\figurewidth] on($(img-14.east) + (-0.1\figurewidth,-0.12\figurewidth)$)  in node [below] at ($(img-14.south) + (0.26\figurewidth,-0.07)$);

        \node [image,right=of img-14,xshift=0.01\figurewidth] (img-15) {\includegraphics[width=\figurewidth,height=0.5\figurewidth]{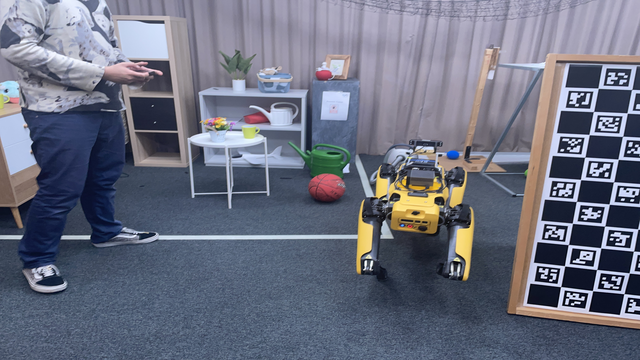}};
        \spy [red,magnification=3,size=0.47\figurewidth] on ($(img-15.west) + (0.1\figurewidth,-0.03\figurewidth)$)  in node [below] at ($(img-15.south) + (-0.26\figurewidth,-0.07)$);  
        \spy [blue,magnification=3,size=0.47\figurewidth] on($(img-15.east) + (-0.1\figurewidth,-0.12\figurewidth)$)  in node [below] at ($(img-15.south) + (0.26\figurewidth,-0.07)$);

        \node[label] (label2) at (img-01.north) {w/o $\beta$};
        \node[label] (label3) at (img-02.north) {$\beta^{s}$ (only)};
        \node[label] (label4) at (img-03.north)  {$\beta^{t}$ (only)};
        \node[label] (label5) at (img-04.north)  {$\beta_{ts}$};
        \node[label] (label6) at (img-05.north)  {GT Image};
        \node[label,rotate=90] (scene1) at ($(img-00.west) + (0,-0.8\figurewidth)$) { Corner};

    \end{tikzpicture}
    \setlength{\belowcaptionskip}{-10pt}
    \caption{\textbf{Distractor Suppression Performance under Different Uncertainty Modeling Strategies.} w/o $\beta$ denotes that no uncertainty is modeled.}
    \label{fig:Distractor}
\end{figure*}
\vspace{-1pt}
\subsection{Ablation Study}
We ablate uncertainty modeling and loss combinations to verify each component’s impact.The quantitative results are summarized in Tab. \ref{Tab3}, with qualitative results shown in Fig. \ref{fig:ablation}.
We first removed the uncertainty modeling, whereupon reconstruction quality dropped markedly, underscoring the mechanism’s critical role in robust scene modeling.
Secondly, when Target-view Uncertainty $\beta^t$ is removed and only Source-view Uncertainty $\beta^s$ is modeled, the model cannot directly perceive transient distractors present in the target view, leading to ineffective suppression and degraded performance.
Conversely, relying solely on Target-view Uncertainty $\beta^t$ is able to localize distractors, yet the model is prone to misjudging inconsistent static structures as distractors, which reduces the precise reconstruction of true geometry.
Moreover, relying on a single loss function cannot fully constrain model optimization. SSIM loss lacks pixel-level precision, often resulting in blurred details or failed renderings. In contrast, MSE loss fails to capture structure-aware similarity and lacks spatial context, limiting the effectiveness of the uncertainty mechanism. Either loss function alone leads to suboptimal model  performance.

\vspace{-1pt}
\subsection{More Comparisons}
\vspace{-5pt}
\paragraph{Robustness Evaluation}
To evaluate the robustness of our method under distractor conditions, we conducted a distractor injection experiment. As shown in Fig. \ref{fig:Robustness}, even with varying levels of distractors in the source views, our method consistently produces renderings with consistent structure. While slightly lower than the distractor-free baseline, the results remain visually acceptable.
This robustness stems from our proposed distractor-aware mthods, which suppresses distractor regions through uncertainty weighting and guides the model to stably learn multi-view aggregation priors, thereby enabling it to remain focused on scene geometry consistency during inference, even in the presence of transient distractors in the source views.

\vspace{-5pt}
\paragraph{Uncertainty Modeling Evaluation}
To fully evaluate different uncertainty modeling strategies, we conduct a qualitative comparison. As show in Fig. \ref{fig:Uncertainty-Modeling} and Fig. \ref{fig:Distractor}, modeling either type of uncertainty alone presents obvious limitations:
Modeling only the Target-view Uncertainty $\beta^t$ allows the model to perceive transient distractors, but it fails to distinguish the sources of reconstruction error, often misjudging inconsistent static structures as distractors, which erroneously suppresses the scene geometry and causes the loss of fine details. 
Furthermore, $\beta^t$ is modeled at a low resolution and then upsampled, which further amplifies the risk of misjudgment.
Source-view Uncertainty $\beta^s$ is modeled at the original resolution during the feed-forward pass, but it cannot directly perceive transient distractors in the target view, causing noticeable artifacts or residual distractions in its rendering results. 
In contrast, Multi-view Uncertainty $\beta_{ts}$ integrates the complementary strengths from both, effectively restraining the failure modes of each individual term, significantly improving the reliability of the supervision signal. Crucially, this mechanism ensures that distractor awareness is driven by multi-view geometric consistency rather than memorizing specific semantic categories; $\beta_t$ is triggered only when objects violate geometric consistency, thereby enabling more accurate geometry reconstruction and distractor suppression.

\vspace{-5pt}
\section{Conclusion and Limitations}
\vspace{-5pt}
We proposes a Multi-view Uncertainty-guided distractor-aware GeNeRF methods, which decouples reconstruction errors by separately modeling Target-view and Source-view Uncertainties, thereby enabling accurate distractor identification and robust suppression.
However, it cannot explicitly identify or remove distractors, relying on a robust supervision mechanism to guide model to focus on structurally stable and reliable regions, which limits performance in highly occlusions cenarios.
In future work, we will explore explicit distractor modeling and integration with dynamic scene understanding to enhance reliability and interpretability.

{
    \small
    \bibliographystyle{ieeenat_fullname}
    \bibliography{main}
}



\end{document}